\RequirePackage{fix-cm}

\documentclass[smallcondensed]{svjour3}     
\smartqed  

\usepackage{latexsym}
\usepackage{mathptmx}      
\usepackage{graphicx}
\usepackage{makecell}
\usepackage{float}
\usepackage{comment}
\usepackage{subfig}
\usepackage[dvipsnames]{xcolor}
\usepackage{algorithm}
\usepackage{algpseudocode}
\usepackage{graphicx}
\usepackage{epsfig}
\usepackage{cite}
\usepackage{tensor}
\usepackage{color}
\usepackage{bm}
\usepackage[cmex10]{amsmath}
\usepackage{amssymb}
\usepackage{amsfonts}
\usepackage{hyperref}
\hypersetup{hypertex=true,
	colorlinks=true,
	linkcolor=blue,
	anchorcolor=blue,
	citecolor=blue}

\graphicspath{{figures/}}
\DeclareGraphicsExtensions{.eps}

\begin{document}

\title{Towards Latent Space Based Manipulation of Elastic Rods using Autoencoder Models and Robust Centerline Extractions
\thanks{J. Qi, G. Ma and Y. Lyu are with the Harbin Institute of Technology, Harbin, 150001, China.}
\thanks{P. Zhou and D. Navarro-Alarcon are with the Hong Kong Polytechnic University, Hung Hom, KLN, Hong Kong. 
Corresponding author: {\texttt{\small  dna@ieee.org}}}
\thanks{Haibo Zhang is with the Beijing Institute of Control Engineering, Beijing, 100190, China.
National Key Laboratory of Science and Technology on Space Intelligent Control, Beijing, 100190, China.}
}

\author{Jiaming Qi  \and
        Guangfu Ma  \and \\
        Peng Zhou  \and 
        Haibo Zhang  \and 
        Yueyong Lyu  \and \\ 
        David Navarro-Alarcon$^*$ 
}

\institute{Jiaming Qi \at
           Harbin Institute of Technology, Harbin, 150001, China. \\
           Tel.: +86-18646086707\\
           \email{qijm\_hit@163.com}           
           \and
           }

\date{Received: date / Accepted: date}

\maketitle
\begin{abstract}
The automatic shape control of deformable objects is a challenging (and currently hot) manipulation problem due to their high-dimensional geometric features and complex physical properties.
In this study, a new methodology to manipulate elastic rods automatically into 2D desired shapes is presented.
An efficient vision-based controller that uses a deep autoencoder network is designed to compute a compact representation of the object's infinite-dimensional shape.
An online algorithm that approximates the sensorimotor mapping between the robot’s configuration and the object’s shape features is used to deal with the latter’s (typically unknown) mechanical properties.
The proposed approach computes the rod’s centerline from raw visual data in real-time by introducing an adaptive algorithm on the basis of a self-organizing network.
Its effectiveness is thoroughly validated with simulations and experiments.

\keywords{Robotics \and Visual Servoing \and Deformable Objects \and Autoencoder \and Self-Organizing Network \and Model Predictive Control}
\end{abstract}

\section{Introduction}\label{intro}
Controlling the shape of soft objects automatically with robot manipulators is highly valuable in many applications, such as food processing \cite{tokumoto2002deformation}, robotic surgery \cite{abolmaesumi2002image}, cable assembly \cite{tang2018framework}, and household works \cite{sun2019general}.
Although great progress has been achieved in recent years, shape control remains an open problem in robotics \cite{navarro2014visual}.
One of the most crucial issues that hamper the implementation of these types of controls is the difficulty to obtain a meaningful and efficient feedback representation of the object’s configuration in real-time.
However, given the intrinsic high-dimensional nature of deformable objects, standard vision-based control algorithms (e.g., based on simple point features) cannot be used as they cannot properly capture the objects’ state. 
In this work, a solution is provided to this problem.

The configuration of rigid objects can be fully described by six degrees of freedom.
However, representing the configuration of soft objects is difficult as they have infinite-dimensional geometric information.
Therefore, a simple and effective feature extractor that can characterize these objects in an efficient (i.e., compact) manner should be designed \cite{cretu2011soft}.
At present, traditional methods are roughly divided into two categories: local and global descriptors.
Local descriptors may use centroids, distances, angles, curvatures \cite{navarro2016automatic} to describe geometric characteristics.
However, these features must be “hard-coded.”
In turn, they can only provide a fixed type of representation.
Global descriptors produce a generic representation of the object’s shape.
An example method under this category is the Point Feature Histogram (PFH) reported in \cite{rusu2008persistent}.
PFH forms a multi-dimensional histogram to represent the overall shape of a soft object.
Subsequent efforts developed PFH into the Fast Point Feature Histograms (FPFH), which reduces computation time \cite{5152473,hu20193}.
A method based on linearly parameterized (truncated) Fourier series was also proposed to represent the object’s contour \cite{navarro2017fourier}.
This parameterization idea was generalized in \cite{qi2020adaptive}, where more shape representations were analyzed and implemented.

Learning-based solutions have received considerable attention due to their potential to learn (in latent space) shape representations of virtually any type of object from data observations only \cite{yeo2005colour}.
Force and position measurements of a three-finger gripper manipulating a soft object were used in \cite{cretu2011soft} as input to a network, which produced and predicted the object’s contour (even for unknown objects).
A coarse-to-fine shape representation was also proposed on the basis of spatial transformer networks, which allowed it to obtain good generalization properties without expensive ground truth observations \cite{yan2020self}.
Growing neural gas was used in \cite{valencia2019toward} to represent deformable shapes.
In \cite{2019Convolutional}, a feature extractor based on the convolutional autoencoder was developed.
This method was used to obtain a low-dimensional latent space from tactile sensing data.

Traditional methods for manipulating soft objects \cite{henrich2000robot} typically need to identify the complex physical properties of the object.
This requirement hinders their application in practice.
Algorithms based on latent spaces present a feasible solution, as they can effectively extract low-dimensional features from a high-dimensional shape space.
For example, convolutional neural networks were used to build the inverse kinematics of a rope \cite{nair2017combining}, learn the physical model of a soft object in the latent space without any prior knowledge of the object \cite{ebert2018visual}, and estimate the rope’s state and combine it with model predictive control \cite{yan2020self}.
However, none of these works has been used to establish an explicitly shaped servo-loop with a latent space representation.
This idea has not been sufficiently explored in the soft manipulation literature.

In the current work, a new solution to the manipulation problem of the elastic rod is proposed.
The novel contributions of this study are listed as follows.

\begin{enumerate}
	\item A centerline extraction algorithm based on self-organizing maps (SOM) is presented for slender elastic rods.
	
	\item A shape feature extraction algorithm is designed using the deep autoencoder network (DAE).
	The proposed method is used to represent the elastic rod with finite-dimensional feature vectors.
	
	\item Detailed simulations and experiments are conducted to validate the effectiveness of the proposed method.
\end{enumerate}
To the best of the authors’ knowledge, this work is the first attempt wherein a shape servo-controller uses DAE to establish an \emph{explicit} shape servo-loop.
The remainder of this study is organized as follows.
The preliminaries are presented in Section \ref{sec2}.
The overall deformation control implementation process is discussed in Section \ref{sec3}.
Various visually servoed deformation tasks of elastic rods are shown in Sections \ref{sec4} and \ref{sec5}.
Conclusions and future work are provided in Section \ref{sec6}.

\section{PRELIMINARIES}\label{sec2}
\emph{Notation.}
Column vectors are denoted with bold small letters $\mathbf{v}$ and matrices with bold capital letters $\mathbf M$. 
Time evolving variables are represented as $\mathbf m_k$, where the subscript $k$ denotes the discrete time instant. 
$\mathbf{E}_n$ is an $n \times n$ identity matrix.

\begin{figure}[t]
	\centering
	\includegraphics[scale=0.92]{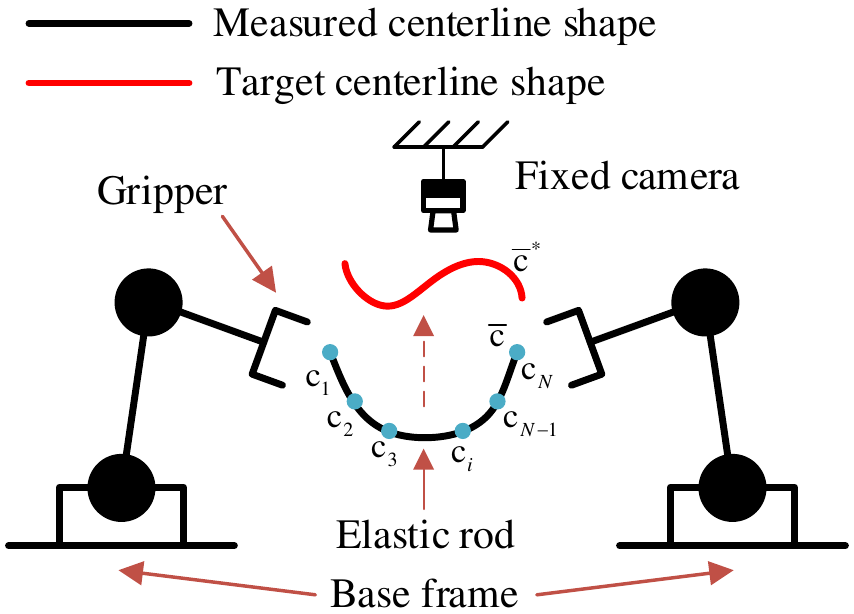}
	\caption{Schematic diagram of the elastic rod shape deformation. The camera is utilized to determine shape feature $\mathbf{s}$ in real time, and within the designed controller the robot automatically deform the real-time shape denoted by $\bar{\mathbf{c}}$ of elastic rods into the target shape $\bar{\mathbf{c}}^*$.}
	\label{fig24}
\end{figure}

The deformation control scheme of a robot manipulating the elastic rod based on visual servoing is investigated.
The following conditions are provided to foster an understanding among readers:
\begin{itemize}
	\item A fixed camera is used to measure the centerline of the elastic rod, namely, eye-to-hand configuration (depicted in Fig. \ref{fig24}). The coordinates obtained are denoted by:
	\begin{equation}
	\label{eq-1}
	\begin{array}{*{20}{c}}
	{\bar{\mathbf{c}} = {{\left[ {\mathbf{c}_1^T, \dots ,\mathbf{c}_N^T} \right]}^T} \in {\mathbb R^{2N}}}&{{\mathbf{c}_i} = {{\left[ {{u_i},{v_i}} \right]}^T} \in {\mathbb R^2}}
	\end{array}
	\end{equation}

	where $N$ represents the number of points that make up the centerline, $u_i$ and $v_i$ represents the coordinates of the $i$th $(i = 1,\cdots,N)$ point in the image frame.
	
	\item Before the experiment, the robot has tightly grasped the elastic rod; that is, object grasping is not the research field of this article. 
	Measurement loss is also not a problem during the manipulation process.
	
	\item The robot supports velocity control mode, which can accurately execute the given desired kinematic commands $\Delta \mathbf{r}_k \in \mathbb R^q$ \cite{siciliano1990kinematic} and satisfy the incremental position motions $\mathbf{r}_{k} = \mathbf{r}_{k-1} + \Delta \mathbf{r}_k$.
	
	\item The robot manipulates the elastic rod at low speeds, so the shape is uniquely determined by elastic potential energy.
	
\end{itemize}

\textbf{Problem Statement}. 
Without any prior physical characteristics of elastic rods, design a model-free vision-based controller which commands the robot to deform the elastic rod into the desired shape in the 2D image space.

\section{Methods}\label{sec3}

\begin{figure}[t]
	\centering
	\includegraphics[scale=0.4]{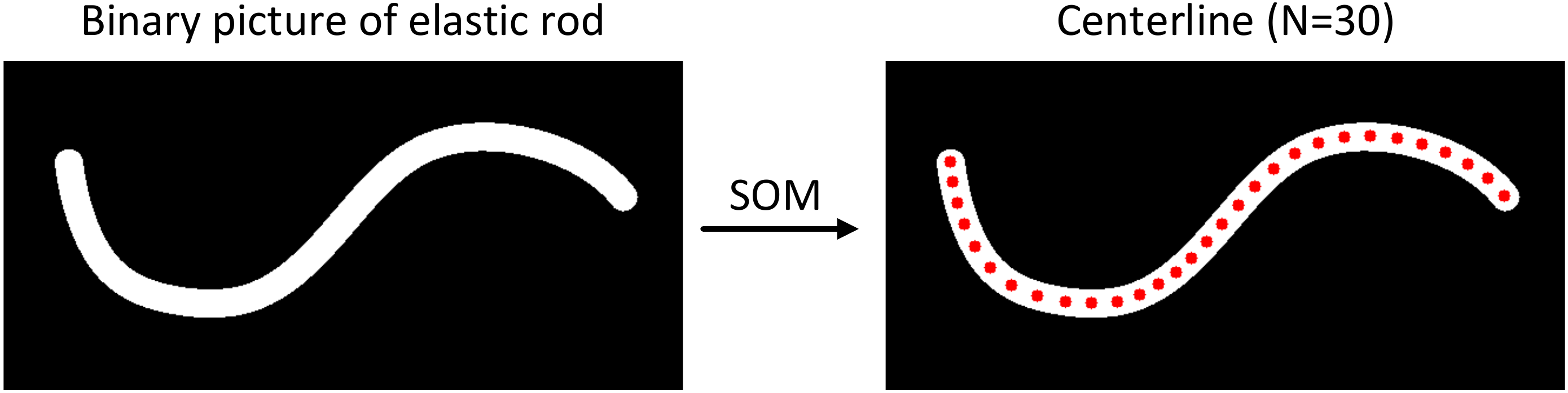}
	\caption{Schematic diagram of SOM-based centerline extraction. 
	The white area in the left side represents the area of elastic rod (clustering area), and the red points in the right side represent the obtained centerline points (clustering points) (in this figure, $N=30$).}
	\label{fig10}
\end{figure}

\subsection{Robust SOM-based Centerline Extraction Algorithm}
\label{sec3c}
Slender elastic rods whose lengths are much larger than their diameters are used as the research object.
Therefore, the centerline describes the shape of the elastic rods.
Given that the centerline generally comprises center-points for elastic rods,
it should be fixed-length, ordered, and equidistant for subsequent feature extraction and controller design.
Although some centerline extraction algorithms are used in the literature, e.g., \emph{OpenCV/thinning}, they cannot meet the above requirements and need pre-processing of data, which will deteriorate the system’s real-time performance.

In this article, SOM is utilized to achieve real-time 2D centerline extraction of elastic rods without artificial marker points.
SOM is a neural network trained in an unsupervised learning manner \cite{kohonen1982self}, which is originally used for dimensionality reduction of high-dimensional data.
Here, it is used as a clustering algorithm.
It generates a fixed number of clustering points from the image data of the elastic rods.
Finally, the centerline is composed of the clustering points.
The input of SOM is the white area where the elastic rod is located in the binary image, as shown in Fig. \ref{fig10}.
The points in the white area are defined by $\bar{\mathbf{m}} = {{\left[ {\mathbf{m}_1^T, \dots ,\mathbf{m}_M^T} \right]}^T} \in \mathbf{R}^{2M}$, $\mathbf{m}_i \in \mathbf{R}^2$ represents coordinates of each point in the image frame, and $M \gg N$.
With the clustering nature of SOM, a fixed-length equidistant centerline can be obtained, namely, $SOM:2M \to 2N$.

\begin{remark}
    The proposed SOM-based centerline extraction is only used in the experiment and not for simulation. 
    The centerline extracted by SOM is not guaranteed to be ordered, so the sorting algorithm \cite{qi2020adaptive} is utilized to reorder the centerline. 
    This process will not take too much time because $N$ is small.
\end{remark}

\begin{figure}[t]
	\centering
	\includegraphics[scale=0.5]{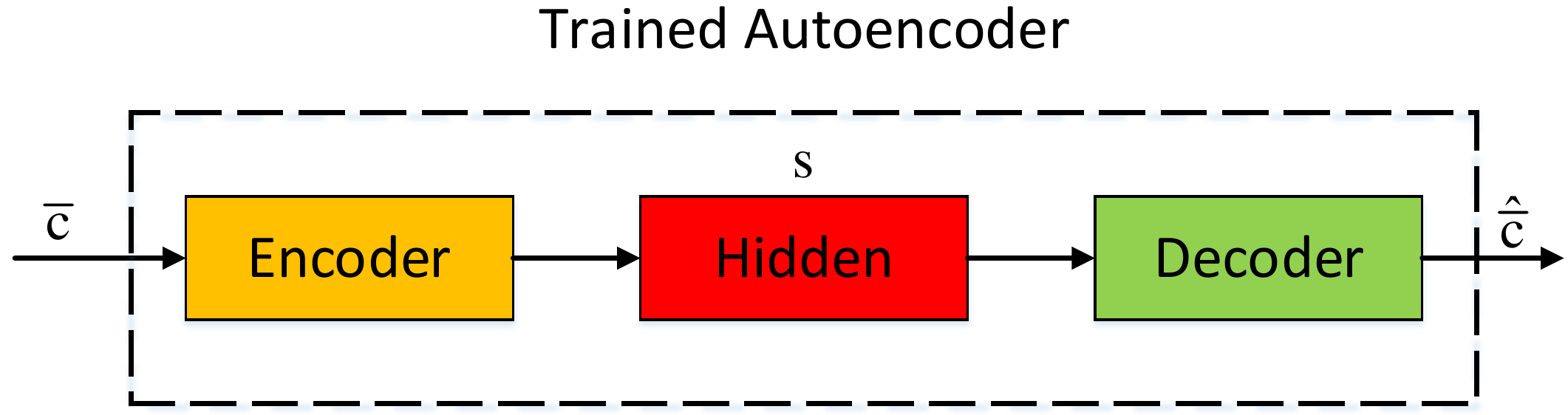}
	\caption{Structure of DAE with the centerline $\overline{\textbf{c}}$ as the input, and $\textbf{s}$ is defined as the shape feature used for deformation Jacobian matrix approximation and controller design.}
	\label{fig4}
\end{figure}

\subsection{Feature Extraction}\label{sec3a}
A controller that can deform the real-time shape $\bar{\mathbf{c}}$ of elastic rods into the target shape $\bar{\mathbf{c}}^*$ can be designed using the centerline extracted by SOM.
However, the centerline cannot be directly inputted into the system.
Given its high dimensionality, it will make the system run slowly and may even cause many adverse effects, e.g., loss of control.
Thus, designing a shape feature extraction algorithm for elastic rods to reduce the feature dimension and represent the centerline effectively is necessary.

In this article, DAE is used to extract shape features $\mathbf{s} \in \mathbb{R}^{p}$ from the high-dimensional centerline $\bar{\mathbf{c}} \in \mathbb{R}^{2N}$.
DAE is an artificial neural network trained in an unsupervised-learning manner, which can automatically learn latent features from unlabeled data \cite{zhou2020lasesom}.
DAE comprises three parts, an Encoder that projects the input into the hidden layer, a hidden layer describing the latent feature $\mathbf{s}$, and a Decoder that reconstructs the latent feature into the original input.
Formally, the centerline $\bar{\mathbf{c}} \in \mathbb{R}^{2N}$ is fed into DAE and mapped to the hidden layer through the nonlinear transformation $\mathbf{s} = \mathbf{f}_{\theta_1} \left( \bar{\mathbf{c}} \right) = sig(\mathbf{W}_1 \bar{\mathbf{c}} + \mathbf{b}_1)$, where parameter set $\theta_1 = \left\{ \mathbf{W}_1, \mathbf{b}_1 \right\}$.
$\mathbf{W}_1$ is a $k \times 2N$ weight matrix, $\mathbf{b}_1$ is a vector of bias and sig is a $sigmoid$ activation function, $s\left( {\bar c} \right) = \frac{1}{{1 + {e^{ - \bar c}}}}$.
The latent feature $\mathbf{s}$ is input into the Decoder to generate
a reconstruction $\hat{\bar{\mathbf{c}}}$ with $2N$ dimensions through the deterministic equation $\hat{\bar{\mathbf{c}}} = \mathbf{g}_{\theta_2} \left( \mathbf{s} \right) = sig(\mathbf{W}_2 \mathbf{s} + \mathbf{b}_2)$, with $\theta_2 = \left\{ \mathbf{W}_2, \mathbf{b}_2 \right\}$.
The parameters of $\theta_1$ and $\theta_2$ of the DAE are designed to minimize the average error of reconstruction, which is defined as:
\begin{equation}
\label{eq15}
\left\{ {\theta _1^*,\theta _2^*} \right\} = \mathop {\arg \min }\limits_{{\theta _1},{\theta _2}} \sum\limits_{k = 1}^N {L\left( {{\mathbf{c}_i},{g_{{\theta _2}}}\left( {{f_{{\theta _1}}}\left( {{\mathbf{c}_i}} \right)} \right)} \right)}
\end{equation}
where $\theta_1^*$ and $\theta_2^*$ are the ideal parameters, and $L$ is usually a mean square error.
Once the Autoencoder is trained, the centerline $\bar{\mathbf{c}}$ is input into the network, and the low-dimensional shape feature $\mathbf{s} \in \mathbb{R}^{p}$ can be obtained through nonlinear transformations $\mathbf{f}_{\theta_1}$.
The workflow of DAE is shown in Fig. \ref{fig4}.

For DAE, the reconstruction output $\hat{\bar{\mathbf{c}}}$ is not the focus, and only the Encoder is utilized to provide the shape feature $\mathbf{s} \in \mathbb{R}^p$ once the DAE is trained. 
At present, DAE has various forms.
In this paper, multilayer perceptron (MLP) is used, given its ability to handle 2D data efficiently.
The size of shape feature dimension $p$ can also be selected due to a trade-off balance.
A small $p$ will improve the system's controllability, e.g., $p<q$.
However, a large $p$ will enhance the representation accuracy of centerlines.
In the simulation and experiment, the effect of various $p$ on the shape representation capabilities is presented.

\subsection{Approximation of the Local Deformation Model}
Given that regular (i.e., mechanically well-behaved) elastic objects are considered, the centerline $\bar{\mathbf{c}}$ is extremely dependent on the robot command $\mathbf{r} \in \mathbb{R}^q$ can be defined as the joint angles or end-effector's pose in this study.
The relationship between $\bar{\mathbf{c}}$ and $\mathbf{r}$ can be represented by the following unknown function \eqref{eq38}.
\begin{equation}
\label{eq38}
\bar{\mathbf{c}} = \mathbf{h} \left(  \mathbf{r} \right)
\end{equation}

Following \eqref{eq38}, the overall kinematics model from robot command $\mathbf{r}$ to shape feature $\bar{\mathbf{c}}$ can be constructed as follows:
\begin{equation}
\label{eq39}
\mathbf{s} = \mathbf{f}_{\theta_1} \left( \mathbf{h} \left(  \mathbf{r} \right) \right)
\end{equation}

Differentiating \eqref{eq39} concerning time variable $t$ yields:
\begin{equation}
\label{eq40}
\dot{\mathbf{s}}  = \mathbf{J} \left( t \right)\dot{\mathbf{r}} 
\end{equation}
where $\mathbf{J} \left( t \right) = \partial \mathbf{s}/\partial \mathbf{r} \in \mathbb R^{p \times q}$ represents a Jacobian-like matrix that 
describes the mapping between the feature change speed $\dot{\mathbf{s}}$ and the velocity command $\dot{\mathbf{r}}$.
The properties of elastic rods are unknown, so the analytical form of $\mathbf{J} (t)$ cannot be obtained.
Discretizing \eqref{eq40} yields the first-order format as follows:
\begin{equation}
\label{eq41}
\mathbf{s}_k = \mathbf{s}_{k-1} + \mathbf{J}_k \Delta \mathbf{r}_k
\end{equation}
where $\Delta \mathbf{r}_k = \mathbf{r}_k - \mathbf{r}_{k-1} \in \mathbf{R}^q$.
The application of DAE as feature extraction is the focus of this study.
Accordingly, the simple Broyden algorithms are used to computes local approximations of $\mathbf J_k$ in real-time.
Define the following differential signal:
\begin{equation}
\label{eq9}
\begin{array}{*{20}{c}}
{{\mathbf{y}_k} = {\mathbf{s}_{k}} - {\mathbf{s}_{k-1}}}&{{\mathbf{u}_k} = \Delta \mathbf{r}_k} = {\mathbf{r}_{k}} - {\mathbf{r}_{k-1}}
\end{array}
\end{equation}

Broyden algorithms are as follows:
\begin{enumerate}
	\item R1 update formula \cite{broyden1965class}:
	\begin{equation}
	\label{eq4}
	\hat{\mathbf{J}}_{k} = \hat{\mathbf{J}}_{k-1} + \frac{{\left( {{\mathbf{y}_k} - \hat{\mathbf{J}}_{k-1}{\mathbf{u}_k}} \right)\mathbf{u}_k^T}}{{\mathbf{u}_k^T{\mathbf{u}_k}}}
	\end{equation}
	This form has a simple structure and fast calculation speed.
	
	\item SR1 update formula \cite{broyden1965class}:
	\begin{equation}
	\label{eq5}
	\hat{\mathbf{J}}_{k} = \hat{\mathbf{J}}_{k-1} + \frac{{\left( {{\mathbf{y}_k} - \hat{\mathbf{J}}_{k-1}{\mathbf{u}_k}} \right){{\left( {{\mathbf{y}_k} - \hat{\mathbf{J}}_{k-1}{\mathbf{u}_k}} \right)}^T}}}{{{\mathbf{u}_k}{{\left( {{\mathbf{y}_k} - \hat{\mathbf{J}}_{k-1}{\mathbf{u}_k}} \right)}^T}}}
	\end{equation}
	The structure of SR1 is similar to R1, but the calculation accuracy is higher.
	
	\item DFP update formula \cite{nocedal1980updating}:
	\begin{align}
	\label{eq7}
	\hat{\mathbf{J}}_{k} &= \hat{\mathbf{J}}_{k-1} + \frac{{\left( {{\mathbf{y}_k} - \hat{\mathbf{J}}_{k-1}{\mathbf{u}_k}} \right)\mathbf{y}_k^T + {\mathbf{y}_k}{{\left( {{\mathbf{y}_k} - \hat{\mathbf{J}}_{k-1}{\mathbf{u}_k}} \right)}^T}}}{{{\mathbf{u}_k}\mathbf{y}_k^T}} \\ \notag
	&- \frac{{\mathbf{y}_k^T{\mathbf{y}_k}}}{{\left\| {{\mathbf{u}_k}\mathbf{y}_k^T} \right\|}}\left( {{\mathbf{y}_k} - \hat{\mathbf{J}}_{k-1}{\mathbf{u}_k}} \right)\mathbf{u}_k^T
	\end{align}
	DFP is a rank two quasi-Newton method, which is efficient for solving nonlinear optimization.
	
	\item BFGS update formula \cite{dennis1974characterization}:
	\begin{equation}
	\label{eq6}
	\hat{\mathbf{J}}_{k} = \hat{\mathbf{J}}_{k-1} - \frac{{\hat{\mathbf{J}}_{k-1}{\mathbf{u}_k}\mathbf{u}_k^T \hat{\mathbf{J}}_{k-1}^T}}{{{\mathbf{u}_k}\mathbf{u}_k^T \hat{\mathbf{J}}_{k-1}^T}} + \frac{{{\mathbf{y}_k}\mathbf{y}_k^T}}{{{\mathbf{u}_k}\mathbf{y}_k^T}}
	\end{equation}
	It is recognized with the best numerical stability.	
\end{enumerate}

When a new data pair $\left(\mathbf{y}_k, \mathbf{u}_k \right)$ enters the system, the deformation Jacobian matrix $\hat{\mathbf{J}}_k$ can be updated with the above estimators.
\begin{remark}
	The robot is assumed to manipulate elastic rods at low speed.
	Thus, the deformation of the elastic rods is relatively slow.
	On the basis, the deformation Jacobian matrix $\mathbf{J}_k$ can be estimated online as the formula \eqref{eq39} is assumed to be smooth.
\end{remark}

\begin{figure}[t]
	\centering
	\includegraphics[scale=0.45]{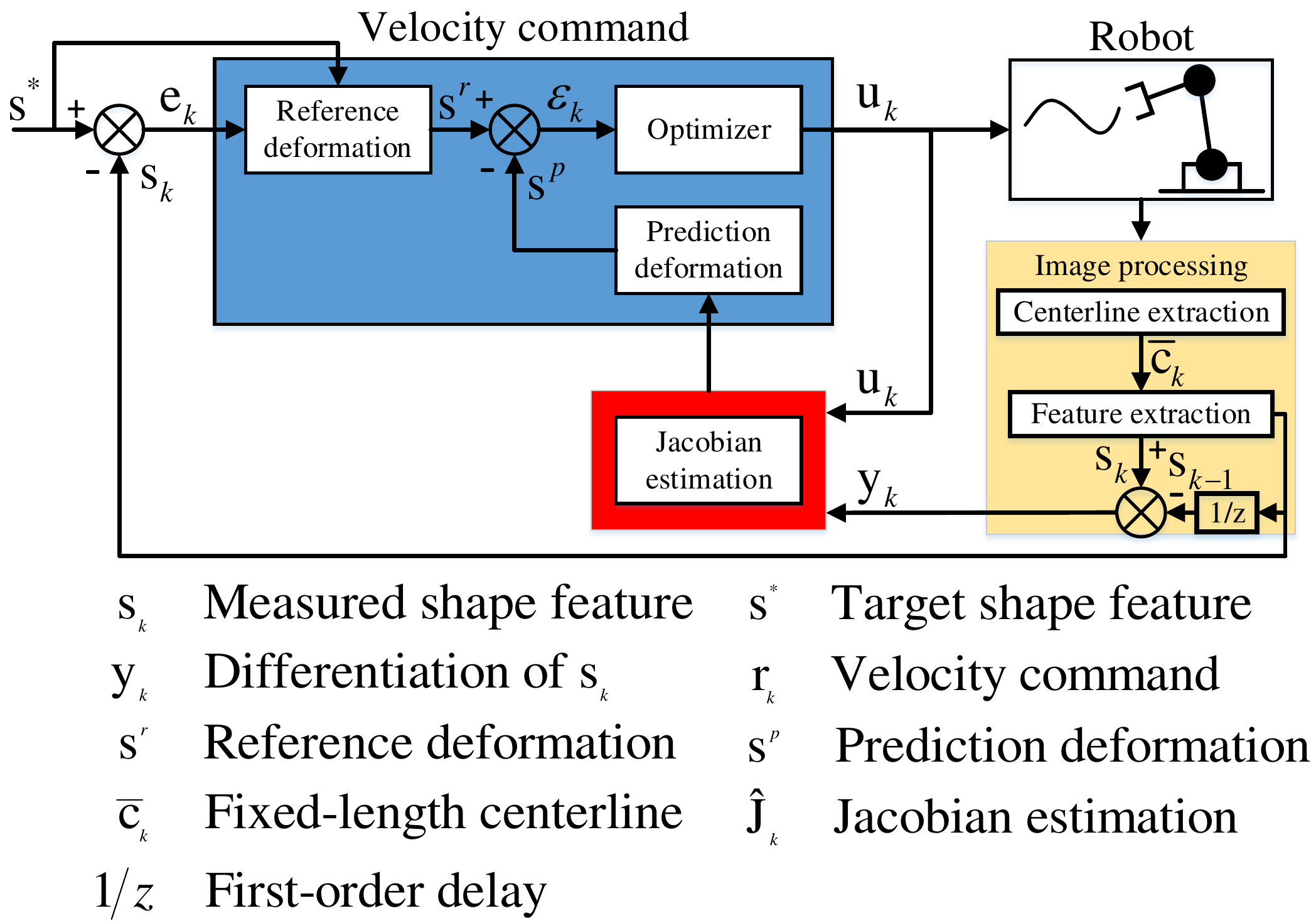}
	\caption{The block diagram of the proposed real-time deformation control strategy.}
	\label{fig23}
\end{figure}

\subsection{Shape Servoing Controller}
At the discrete-time instant $k$, the deformation Jacobian matrix $\hat{\mathbf{J}}_k$ has been assumed to be exactly approximated by \eqref{eq4}\eqref{eq5}\eqref{eq7}\eqref{eq6}, so that the shape-motion difference model is satisfied:
\begin{equation}
\label{eq35}
{\mathbf{s}_k} = {\mathbf{s}_{k - 1}} + {\hat{\mathbf{J}}_k} \cdot {\mathbf{u}_k}
\end{equation}

A model predictive controller \cite{ouyang2018robust} is utilized to minimize the shape deformation error $\mathbf{e}_k = {\mathbf{s}^* - \mathbf{s}_k}$ between the measured feature $\mathbf{s}_k$ and a constant target feature $\mathbf{s}^*$. 
With the estimated deformation Jacobian matrix $\hat{\mathbf{J}}_k$ and \eqref{eq35}, the predicted deformation output at time $k+w$ is shown below:
\begin{equation}
\label{eq45}
s_{k + w}^p = {s_k} + \hat{\mathbf{J}}_k \cdot {\mathbf{u}_{k + w}}
\end{equation}
where $w \in \left[ {0,H} \right]$ represents the length of prediction horizon, and $\mathbf{u}_{k+w} =\mathbf{r}_{k+w} - \mathbf{r}_k$.
The reference deformation trajectory at time $k+w$ is calculated to ensure smooth deformation of elastic rods and the estimation accuracy of deformation Jacobian matrix as follows:
\begin{equation}
\mathbf{s}_{k + w}^r = {\mathbf{s}^*} - {e^{- \rho w}} \cdot {\mathbf{e}_k}
\end{equation}
where $\rho$ is a positive constant.
Error $\varepsilon$ between the reference and the prediction deformation at instant $k + w$ is defined as follows:
\begin{equation}
{\varepsilon _{k + w}} = \mathbf{s}_{k + w}^r - \mathbf{s}_{k + w}^p = \left( {1 - {e^{ - \rho w}}} \right){\mathbf{e}_k} - {\hat{\mathbf{J}} _k}{\mathbf{u}_{k + w}}
\end{equation}

Velocity command $\mathbf{u}_k$ is assumed to remain constant from $k$ to $k + w$ and can be calculated by minimizing $\varepsilon$ from $k$ to $k + H$, as shown below:
\begin{equation}
\label{eq32}
\min \frac{1}{2}\left( {\sum\limits_{w = 0}^H {{\alpha ^w}{{\left\| {\left( {1 - {e^{ - \rho w}}} \right){\mathbf{e}_k} - w{\hat{\mathbf{J}}_k}{\mathbf{u}_k}} \right\|}^2}}  + \mathbf{u}_k^T \mathbf{Q}{\mathbf{u}_k}} \right)
\end{equation}
where $0 < \alpha  \le 1$, and $\mathbf{Q}$ is a symmetric and positive definite matrix used to adjust the command $\mathbf{u}_k$.
When the command $\mathbf{u}_k$ is too large, it will cause the robot to move too fast and the manipulated object will oscillate.
In turn, the estimation accuracy of the deformation Jacobian matrix will be affected.
Taking derivative of \eqref{eq32} with respect to $\mathbf{u}_k$, the gradient $\nabla $ is calculated as follows:
\begin{equation}
\label{eq33}
\nabla  = \sum\limits_{w = 0}^H { - w{\alpha ^w}\hat{\mathbf{J}}_k^T\left( {\left( {1 - {\beta ^w}} \right){\mathbf{e}_k} - w{\hat{\mathbf{J}}_k}{\mathbf{u}_k}} \right)} + \mathbf{Q}{\mathbf{u}_k}
\end{equation}
where $\beta = e^{-\rho}$.
By setting $\nabla = 0$, the velocity command $\mathbf{u}_k$ is derived:
\begin{align}
\label{eq34}
\mathbf{u}_k &= {\left( {a\hat{\mathbf{J}}_k + {\hat{\mathbf{J}}^{T + }}_k\mathbf{Q}} \right)^ + }\left( {b - c} \right)\mathbf{e}_k \notag \\ 
a &= \left( {{H^2}{\alpha ^H} - 2b} \right)/\ln \alpha \notag \\ 
b &= \left( {H{\alpha ^H}\ln \alpha  - {\alpha ^H} + 1} \right)/{\ln ^2}\alpha \\
c &= \left( {H{{\left( {\alpha \beta } \right)}^H}\ln \left( {\alpha \beta } \right) - {{\left( {\alpha \beta } \right)}^H} + 1} \right)/{\ln ^2}\left( {\alpha \beta } \right) \notag
\end{align}

Thus, at each time instant, the incremental position command is calculated as follows:
\begin{equation}
\mathbf{r}_k = \mathbf{r}_{k-1} + \mathbf{u}_k
\end{equation}

Following \eqref{eq35}, it yields:
\begin{equation}
\label{eq36}
{\mathbf{e}_k} - {\mathbf{e}_{k - 1}} =  - {\hat{\mathbf{J}}_k}\Delta {\mathbf{r}_k}
\end{equation}

$\hat{\mathbf{J}}_k$ is assumed to be a full column rank, and substituting \eqref{eq34} into \eqref{eq36} yields:
\begin{equation}
\label{eq37}
\left( {a\mathbf{E}_n + {\hat{\mathbf{J}}_k^{T + }} \mathbf{Q}  \hat{\mathbf{J}}_k^ + } \right)\left( {{\mathbf{e}_k} - {\mathbf{e}_{k - 1}}} \right) + \left( {b - c} \right){\mathbf{e}_k} = 0 
\end{equation}

As $a > 0, b-c > 0$, and ${\hat{\mathbf{J}}_k^{T + }} \mathbf{Q}  \hat{\mathbf{J}}_k^ + $ is a positive-definite matrix, the error $\mathbf{e}_k$ asymptotically converges to error, namely, $\mathop {\lim }\limits_{t \to \infty } {\mathbf{s}_k} = \mathbf{s}_k^*$.
However, when the reachability of the desired goal $\mathbf{s}_k^*$ is not satisfied, $\hat{\mathbf{J}}_k$ may not be a column full-rank matrix.
The feedback error $\| \mathbf{e}_k \|$ may only converge to the neighborhood near the origin.
For such under-actuated visual servo control tasks, guaranteeing the global asymptotic convergence is challenging \cite{hutchinson2006visual}.
The block diagram of the proposed real-time deformation control strategy is shown in Fig. \ref{fig23}.

\begin{remark}
The velocity controller \eqref{eq34} and deformation Jacobian estimators \eqref{eq4}\eqref{eq5}\eqref{eq7}\eqref{eq6} only require visual feedback data without any additional sensors, prior knowledge of the system model, and the requirement to calibrate the camera.
\end{remark}

\begin{figure}[t]
	\centering
	\includegraphics[scale=0.28]{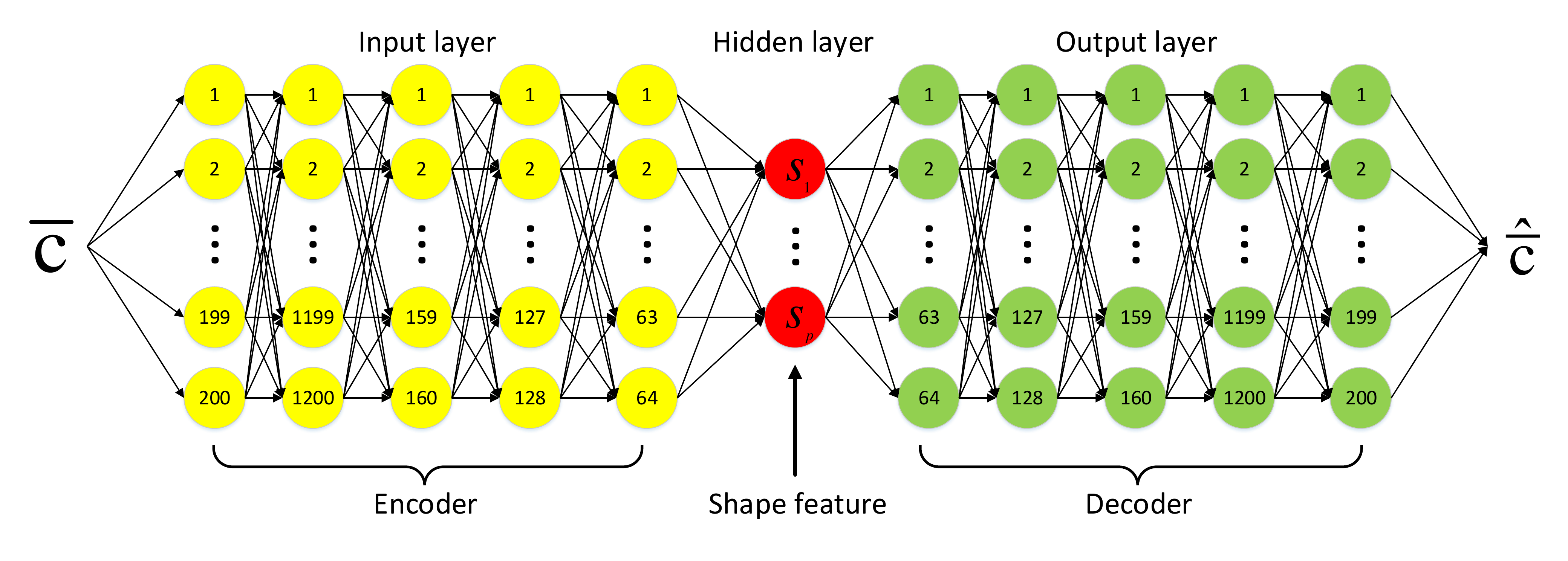}
	\caption{Structure of DAE comprised of MLPs as the basic blocks of Encoder and Decoder.
	The centerline $\bar{\mathbf{c}}$ is fed into the trained DAE to generate shape feature denoted by $\mathbf{s}$.}
	\label{fig30}
\end{figure}

\section{SIMULATION RESULTS}\label{sec4}
The following case is considered: one end of an elastic rod is rigidly grasped by a planar robot (2DOF) and the other end is static. 
For brevity, the robot is not shown in the figures.
The cable simulator is simulated as in \cite{wakamatsu1995modeling} by using the minimum energy principle \cite{hamill2014student}, and publicly available at \url{https://github.com/q546163199/shape_deformation/tree/master/python/package/shape_simulator}.
All numerical simulations are implemented in Python.

\begin{figure}
	\centering
	\includegraphics[scale=0.35]{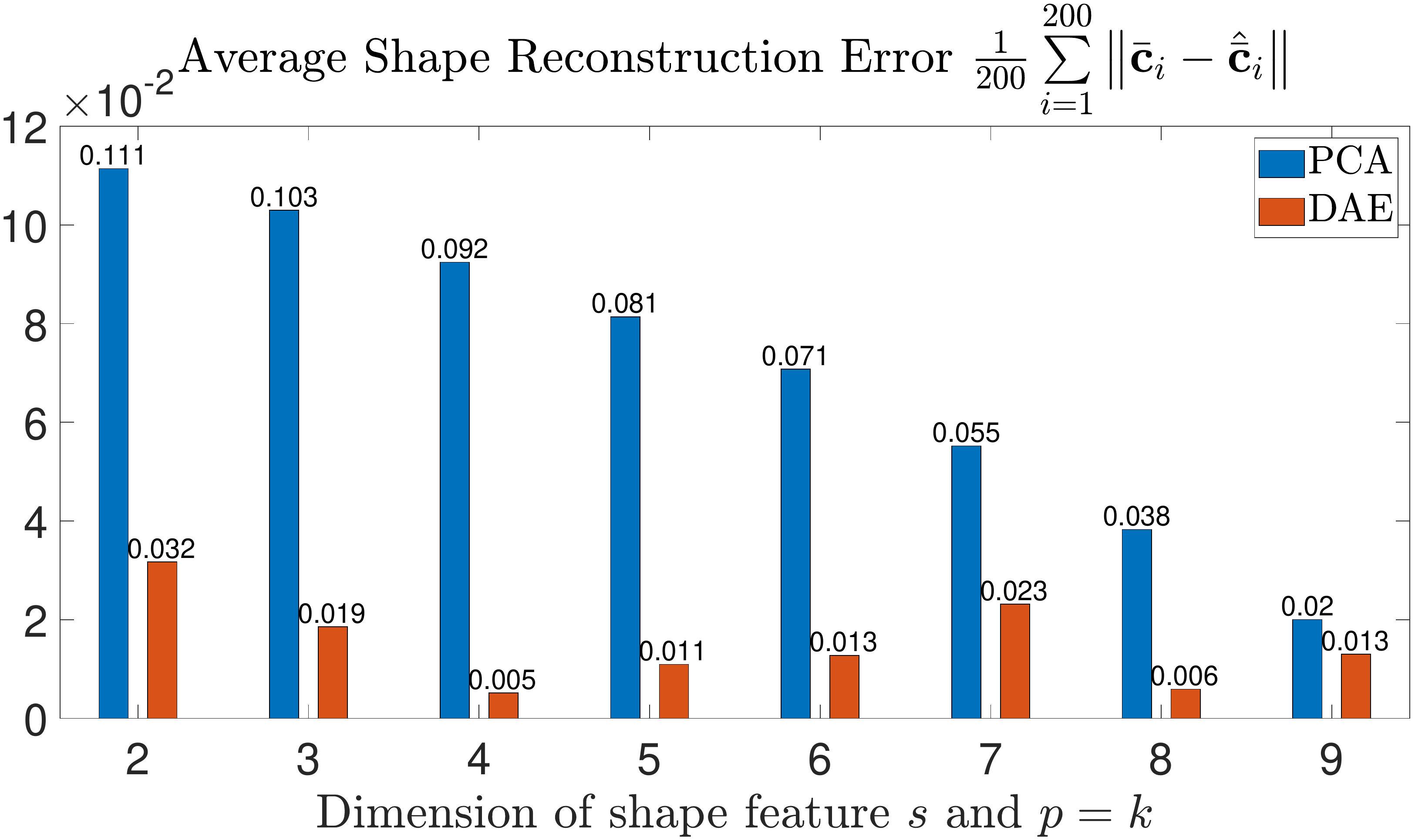}
	\caption{Average shape reconstruction error comparison between DAE and PCA among 200 shape sets in the simulation.}
	\label{fig1}
\end{figure}

\subsection{Feature Extraction Comparison}\label{sec4a}
In this section, 40,000 samples ($N=100$) utilized to train the DAE are generated by randomly moving the robot.
As previously mentioned in Section \ref{sec3a}, DAE comprises MLPs, as shown in Fig. \ref{fig30}.
DAE is implemented on PyTorch and trained by adopting an ADAM optimizer with an initial learning rate of 0.001 and a batch size of 500.
RELU activation functions are adopted in the Encoder and Decoder.

In Fig. \ref{fig1}, the reconstruction error between the simulated shape $\bar{\mathbf{c}}$ and the shape $\hat{\bar{\mathbf{c}}}$ obtained from DAE or PCA is denoted by $\left \| \bar{\mathbf{c}} - \hat{\bar{\mathbf{c}}} \right \|$.
$k$ and $p$ determine the dimension of shape feature $\mathbf{s}$ obtained from PCA \cite{zhu2020vision} and DAE, respectively.
For the fairness of competition, $k$ is set to be equal to $p$ for validating the feature extraction and reconstruction performance of PCA and DAE under the same shape feature dimension.
The result shows that in each case, the reconstruction performance of DAE is better than that of PCA.
For DAE, the results show that $p=4$ has the best reconstruction performance, followed by $p=6$ and $p=2$.
This finding indicates that $p$ is too low to represent the elastic rod fully.
Considering the trade balance of system controllability and shape representation performance, DAE with ${p}=4$ is used in the following sections.

\subsection{Validation of the Jacobian Estimation}\label{sec4b}
\begin{figure}
	\centering
	\subfloat[]{\includegraphics[scale=0.2]{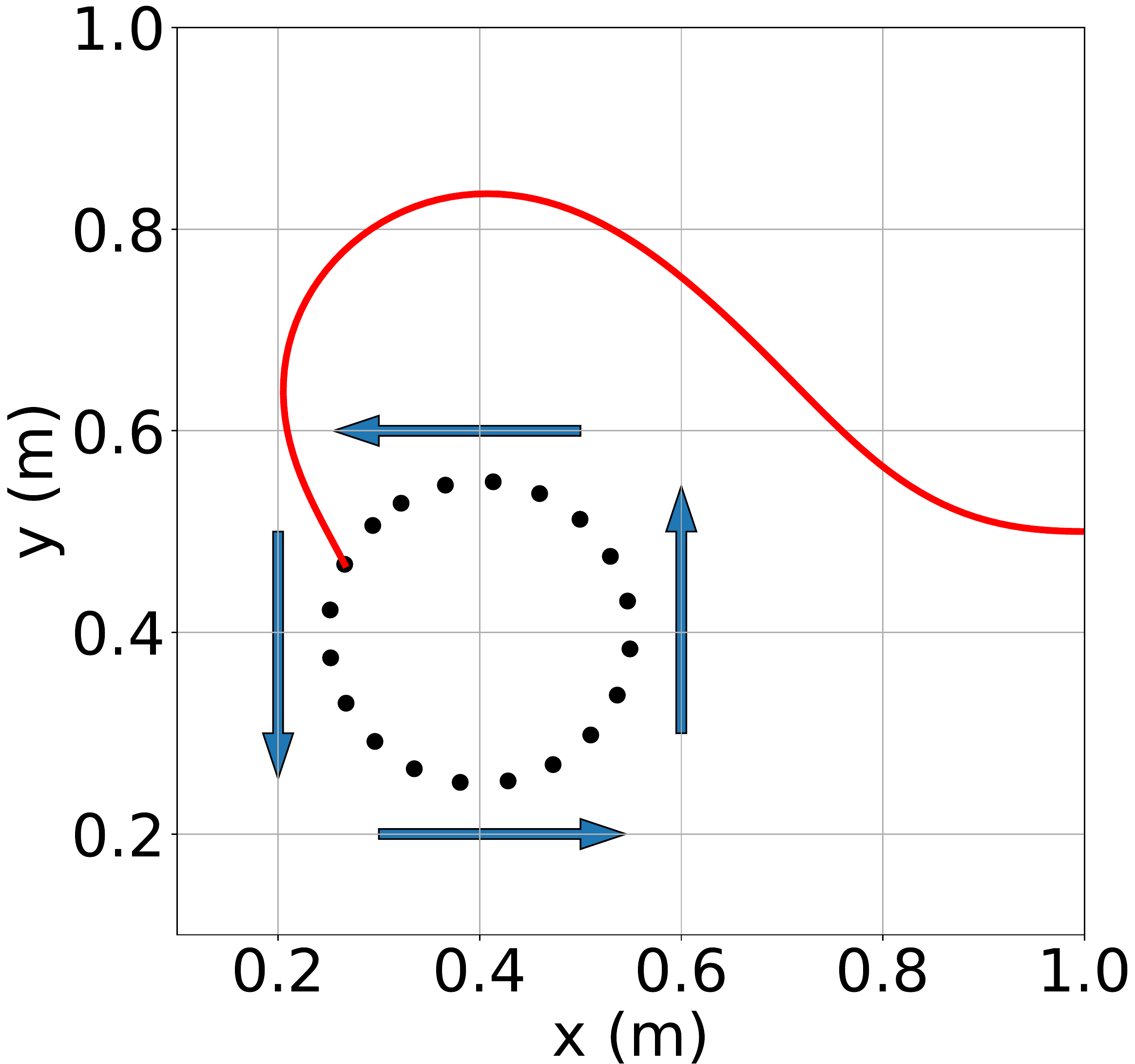}\label{fig21a}}
	\subfloat[]{\includegraphics[scale=0.2]{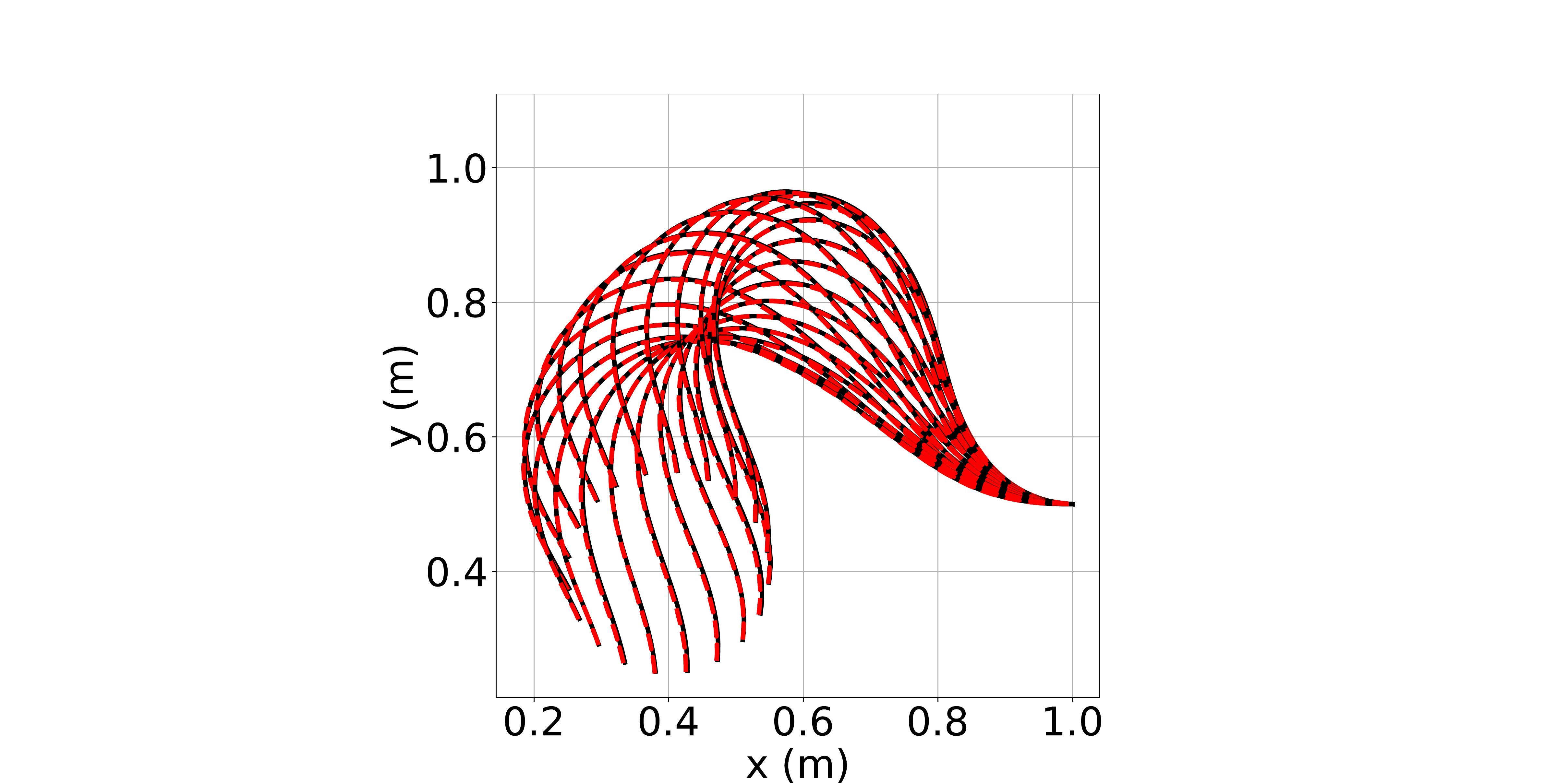}\label{fig21b}}
	
	\caption{Deformation Jacobian matrix $\hat{\mathbf{J}}_k$ validation framework. 
	(a) Motion trajectory of robot's end-effector.
	(b) Comparison between the simulated cable profile (black solid line) and its reconstruction shape obtained by DAE (red dashed line) with $p=4$.}
	\label{fig5}
\end{figure}

In this section, four deformation Jacobian estimators, namely, R1, SR1, DFP, and BFGS, are considered, and their effectiveness is evaluated.
The planar robot grasps one end of the simulated rod and conducts a counterclockwise circular motion with center (0.4, 0.4), as shown in Fig. \ref{fig21a}.
Different from the simple initialization of $\hat{\mathbf{J}}_0$, e.g., the identity matrix, the robot moves in the initial sampling area (the motions are ensured not to be collinear and close to the starting point) to initialize $\hat{\mathbf{J}}_0$.
The initialization accuracy of deformation Jacobian matrix is improved using this method.
This method can also reduce the possibility of singular problems under the deformation Jacobian matrix in the manipulation process.
In turn, the safety of operations is enhanced.
Two error criteria \eqref{eq13} are utilized to compare each deformation Jacobian estimator qualitatively.
\begin{equation}
\label{eq13}
\begin{array}{*{20}{c}}
{{T_1} = \left\| {{\mathbf{s}_k} - {{\hat{\mathbf{s}}}_k}} \right\|}&
{{T_2} = \left\| {\Delta {\mathbf{s}_k} - {{\hat{\mathbf{J}}}_k}\Delta {\mathbf{r}_k}} \right\|}
\end{array}
\end{equation}
where $\mathbf{s}_k$ is feedback shape feature generated by DAE, $\hat{\mathbf{s}}_k$ is calculated by \eqref{eq14}.
\begin{equation}
\label{eq14}
\hat{\mathbf{s}}_k = \hat{\mathbf{s}}_{k-1} + \hat{\mathbf{J}}_k \Delta \mathbf{r}_k
\end{equation}

\begin{figure}
	\centering
	\includegraphics[scale=0.3]{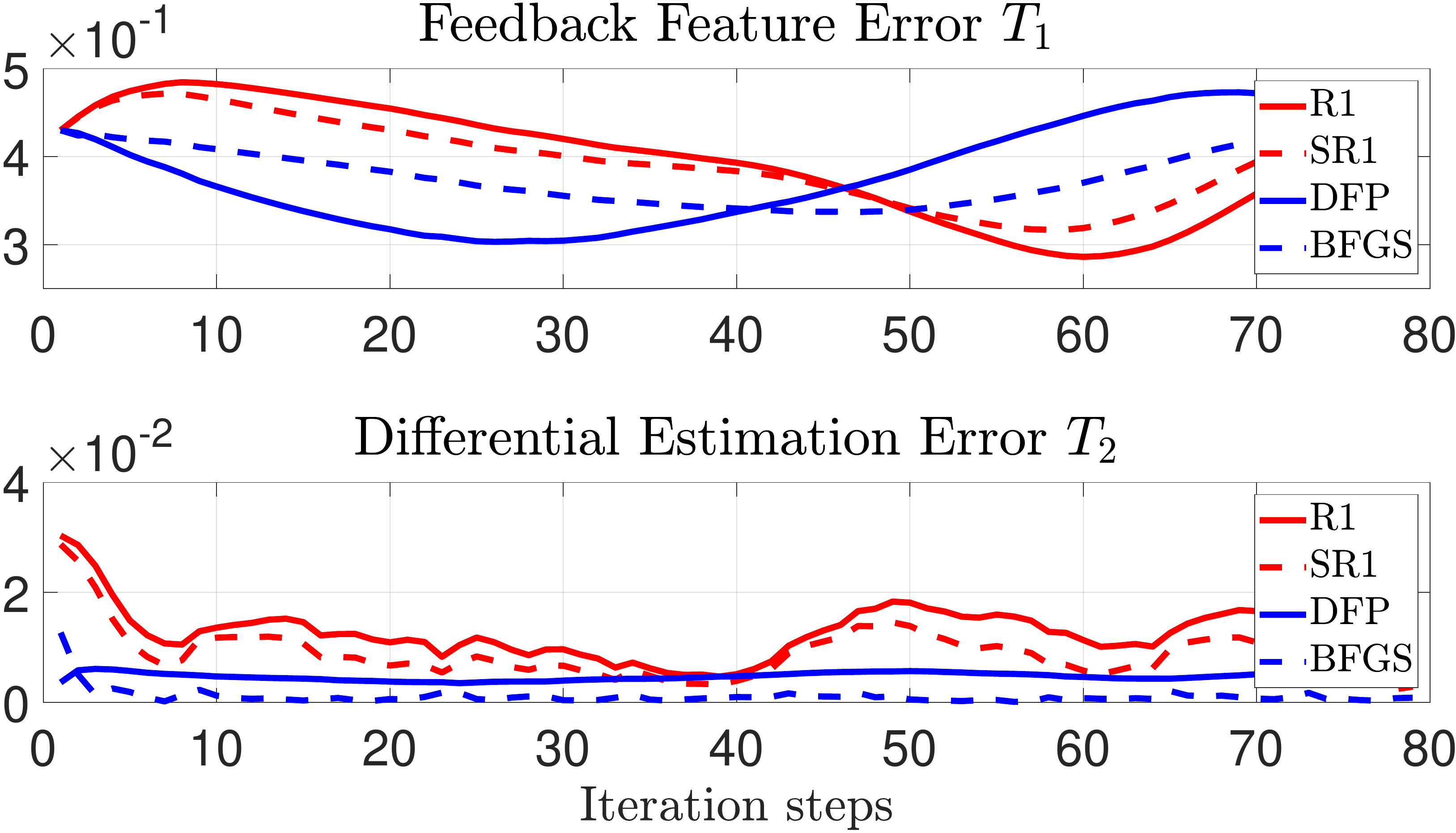}
	\caption{Profiles of the criteria $T_1$ and $T_2$ that are computed along the circular trajectory around the center (0.4, 0.4).}
	\label{fig2}
\end{figure}

The deformation Jacobian estimators and the shape reconstruction accuracy of DAE are verified, as depicted in Fig. \ref{fig21b}.
The results show that the shape reconstruction accuracy of DAE ($p=4$) is well, proving the effectiveness of DAE in the shape representation.
The plots of $T_1$ and $T_2$ during the circular motion are demonstrated in
Fig. \ref{fig2}.
For the $T_1$ curve, all the four deformation Jacobian estimators can accurately update the deformation Jacobian matrix $\hat{\mathbf{J}}_k$, and the average error of BFGS is the smallest.
For $T_2$, BFGS has no apparent fluctuations, and the estimation accuracy is the best.
DFP is second-best, and R1 and SR1 share a similar pattern, consistent with the theoretical analysis.
The above analyses prove that, when starting deformation, BFGS can calibrate and update the deformation Jacobian matrix in time to identify the pseudo-physical parameters of the elastic rods. 
Specifically, BFGS can estimate the change direction of shape feature $\mathbf{s}$ in the latent space.

\subsection{Manipulation of Elastic Rods}\label{sec4c}
\begin{figure}
	\centering
	\subfloat[R1 result]{\includegraphics[scale=0.2]{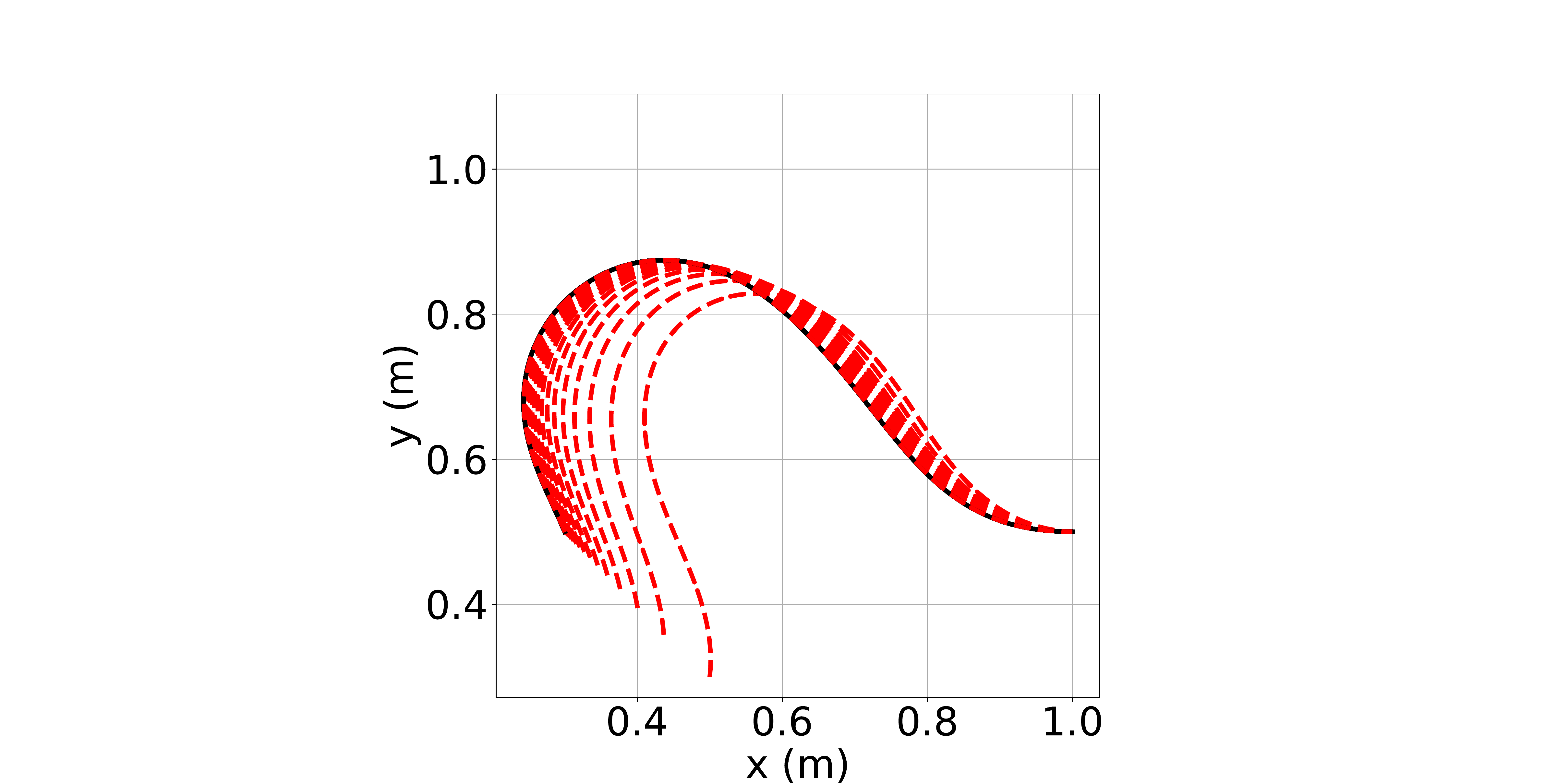}}
	\subfloat[SR1 result]{\includegraphics[scale=0.2]{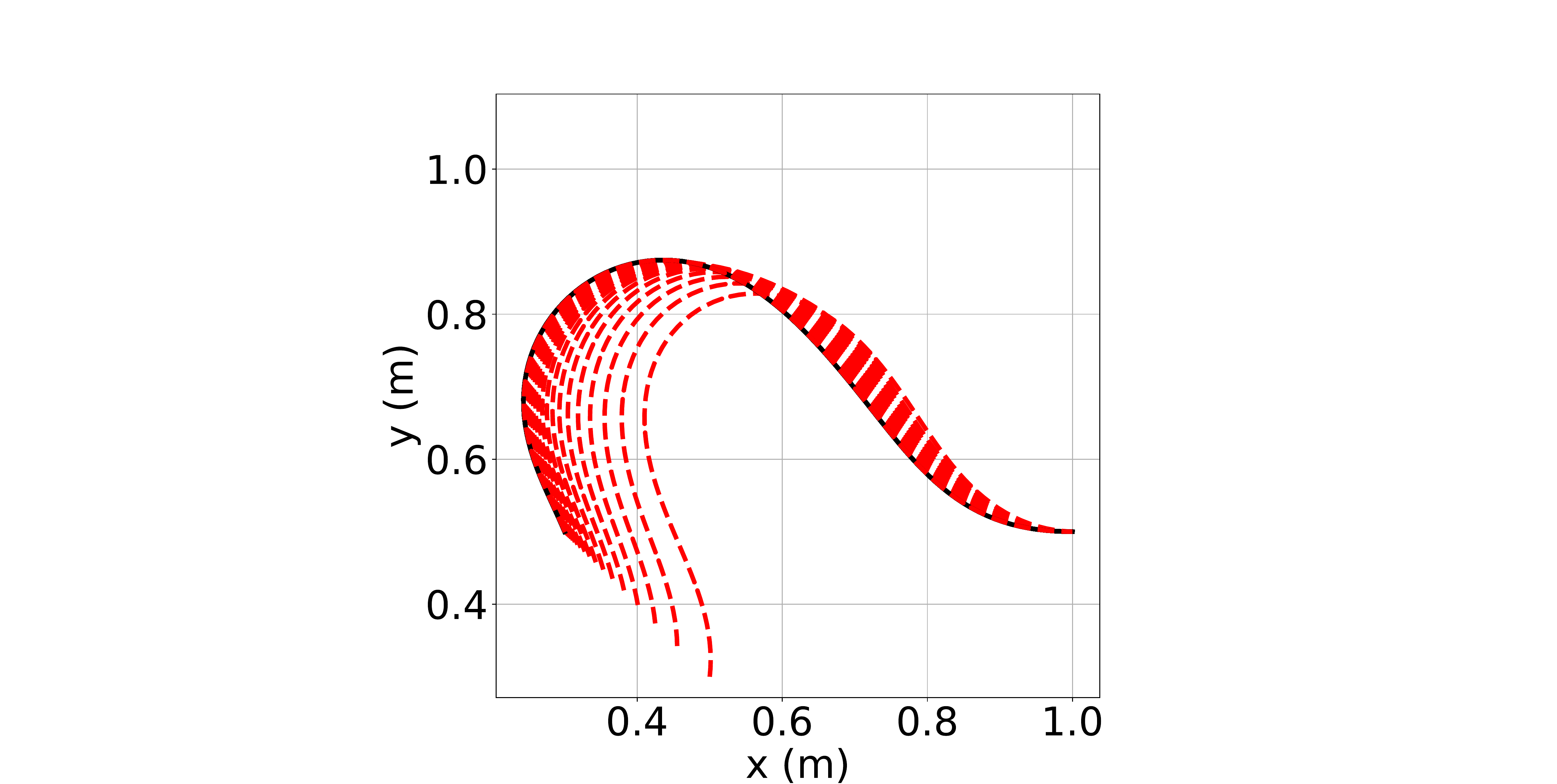}}
	
	\subfloat[DFP result]{\includegraphics[scale=0.2]{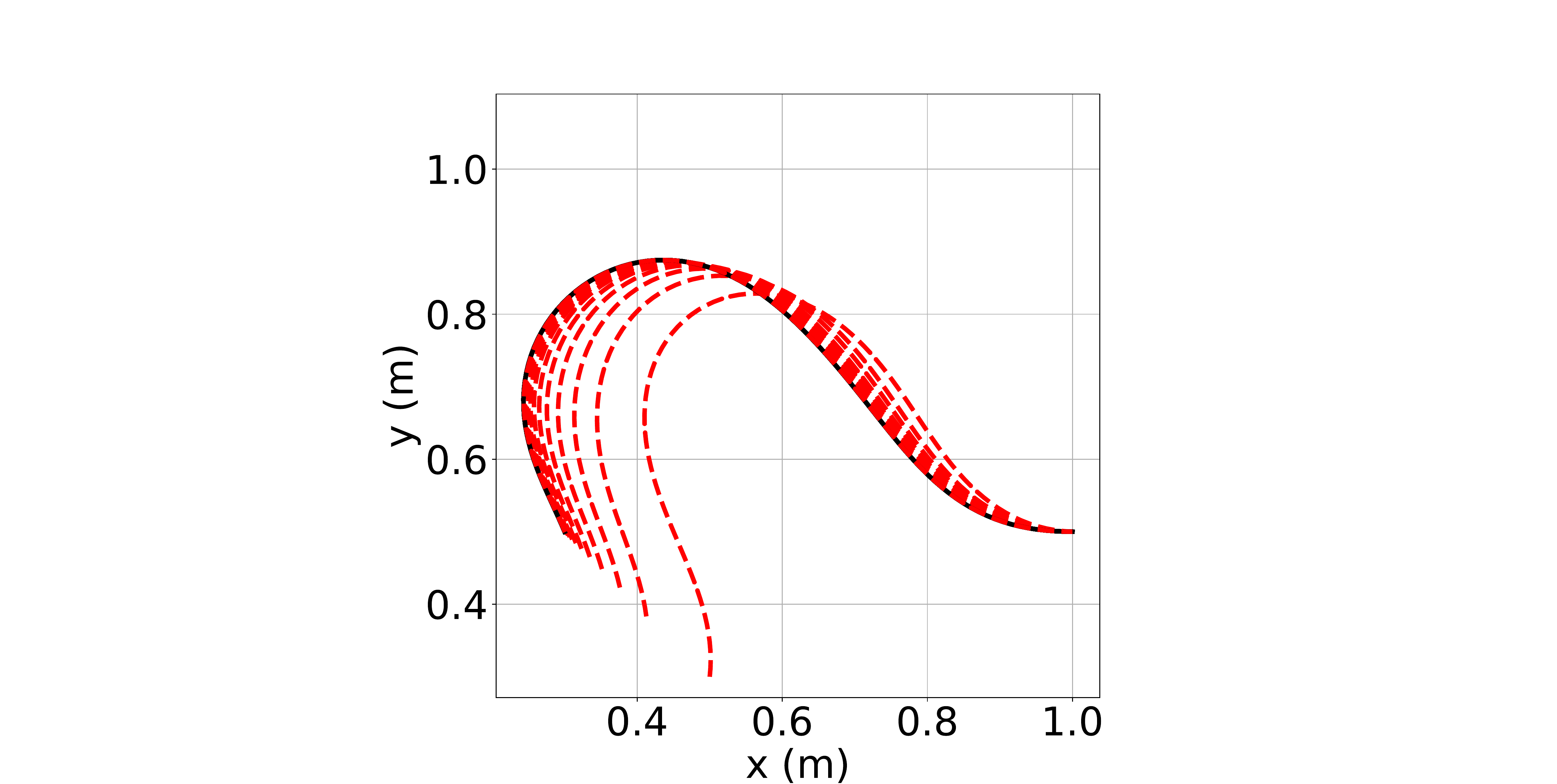}}
	\subfloat[BFGS result]{\includegraphics[scale=0.2]{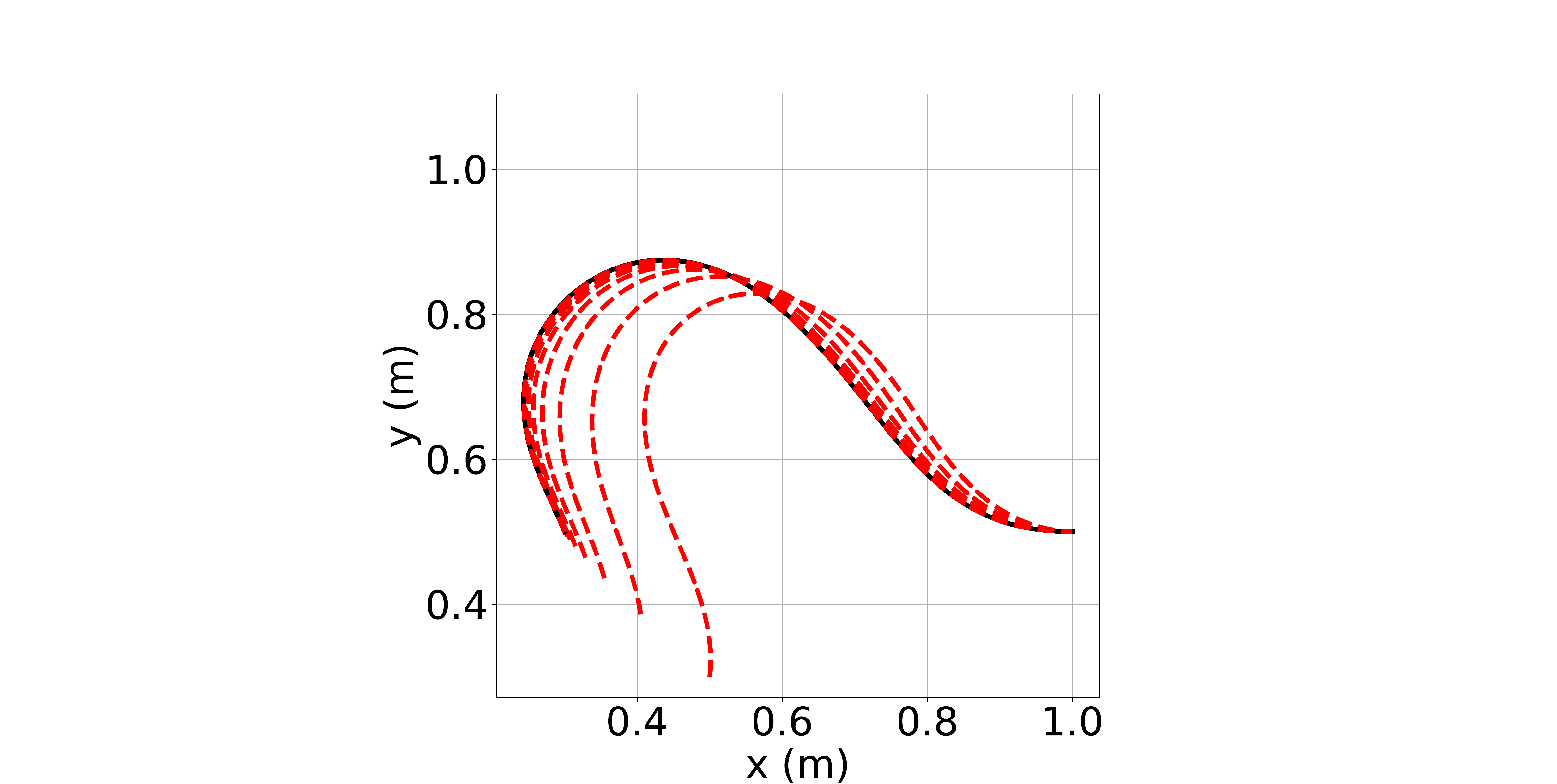}}
	\caption{Profiles of of the shape deformation simulation among R1, SR1, DFP and BFGS (red dashed curves represent the initial and transitional trajectories, and the black solid curve represents the target shape $\bar{\mathbf{c}}^*$ with shape feature $\mathbf{s}^*$).}
	\label{fig17}
\end{figure}

\begin{figure}
	\centering
	\includegraphics[scale=0.3]{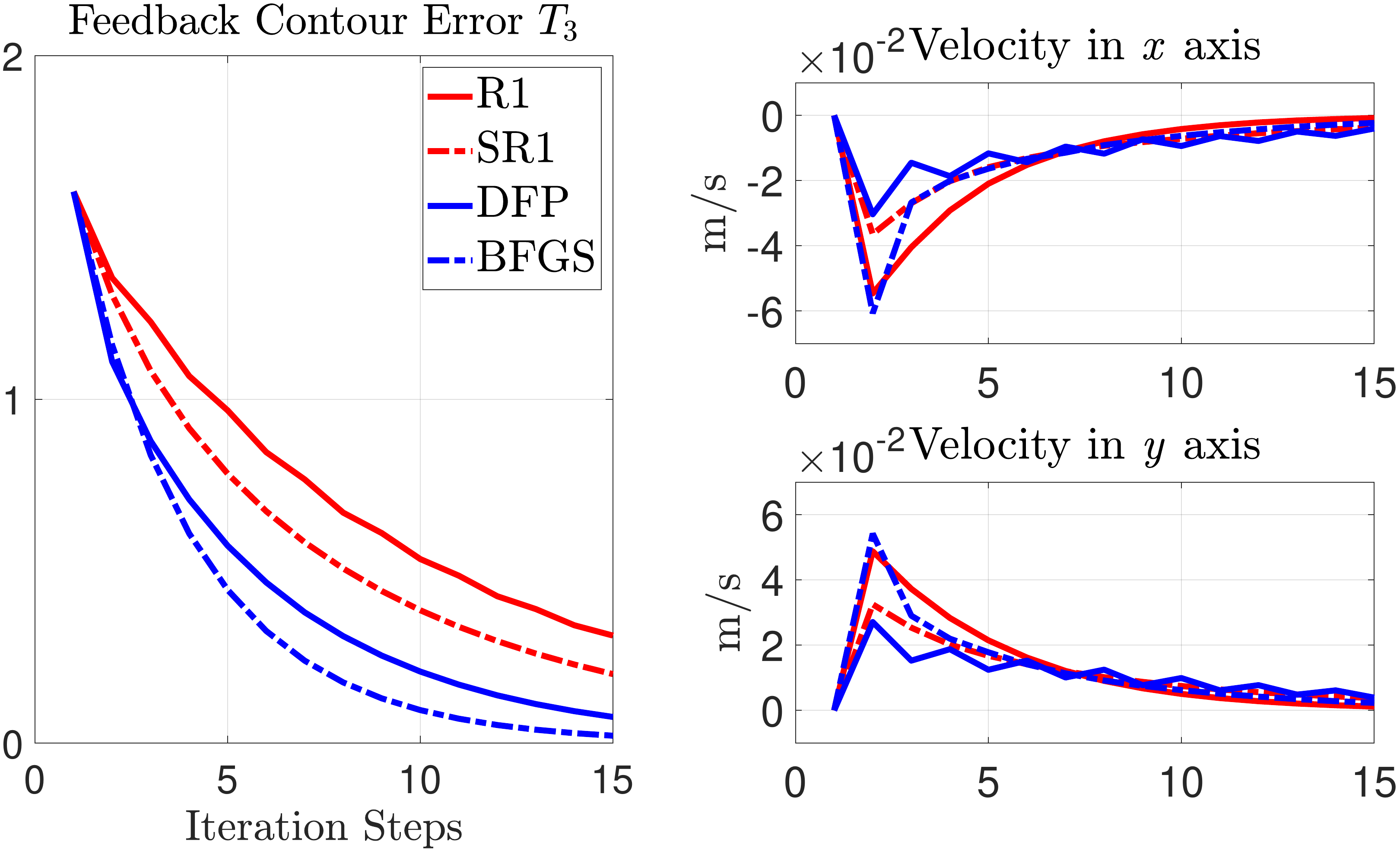}
	\caption{Profiles of the criterion $T_3$ and velocity command $\Delta \mathbf{r}_k$ among R1, SR1, DFP and BFGS within manipulation.}
	\label{fig6}
\end{figure}

In this section, the robot is commanded by the velocity controller \eqref{eq34} to deform the elastic rods into the desired constant shape $\bar{\mathbf{c}}^*$, corresponding to $\mathbf{s}^*$.
The error criterion \eqref{eq16} is utilized to assess the deformation performance. 
\begin{equation}
\label{eq16}
{T_3} = \left\| {{{\bar{\mathbf{c}}}_k} - {{\bar{\mathbf{c}}}}_k^*} \right\|
\end{equation}
The progress of the cable deformation under the velocity command \eqref{eq34} on the basis of R1, SR1, DFP and BFGS is depicted in Fig. \ref{fig17}.
The curve of $T_3$ and the velocity command $\Delta \mathbf{r}_k$ are shown in Fig \ref{fig6}, and the detailed time comparison is provided in Table \ref{table2}.
Both figures show that BFGS is the best method with the shortest convergence time and smallest deformation error, followed by DFP, and the effects of R1 and SR1 are similar.

\begin{table}
	\caption{Results among R1, SR1, DFP and BFGS}
	\label{table2}
	\begin{tabular}{lllll}
		\hline\noalign{\smallskip}
		 & R1 & SR1 & DFP & BFGS \\
		\noalign{\smallskip}\hline\noalign{\smallskip}
		Steps   		& 58 	& 48 	& 32 	& 22		\\
		Time (second)   & 33.64 & 27.84 & 18.56 & 12.76 	\\
		\noalign{\smallskip}\hline
	\end{tabular}
\end{table}

\begin{figure}
	\centering
	\includegraphics[scale=0.7]{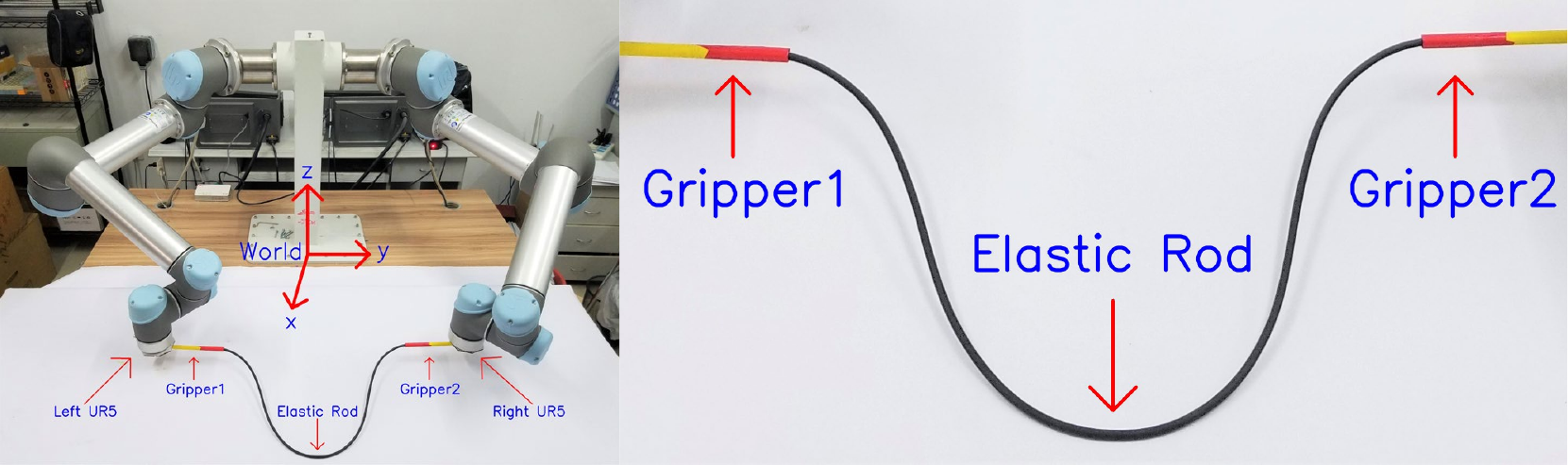}
	\caption{Experimental setup comprised of two UR5 which support velocity control mode.}
	\label{fig20}
\end{figure}

\section{EXPERIMENTAL RESULTS}
\label{sec5}
Various experiments with two UR5 that support velocity control mode are conducted, as shown in Fig. \ref{fig20}.
$\Delta \mathbf{r} = [\Delta \mathbf{r}_{1}^T, \Delta \mathbf{r}_{2}^T]^T \in \mathbb{R}^6$.
$\Delta r_{i1}$ and $\Delta r_{i2}$, $i=1,2$ represent the linear velocity of end-effector along x-axis and y-axis of each UR5 in the world frame.
$\Delta r_{i3}$, $i=1,2$ represents the angular velocity of the sixth joint of each UR5 along the direction parallel to the z axis in the world frame.
A Logitech C270 camera is used to capture the rod's image and combined with OpenCV to process on the Linux PC at 30 fps.
The deformation trajectories display once every two frames to compare the convergence effects of each algorithm.
An experimental video can be downloaded here \url{https://github.com/q546163199/experiment_video/raw/master/paper2/video.mp4}.

\begin{table}[h]
	\caption{Comparison results among three centerline extraction algorithms with $N=50$}
	\label{table4}
	\begin{tabular}{llll}
		\hline\noalign{\smallskip}
		& Reference \cite{qi2020adaptive} & CL \cite{park2000centroid} & SOM \\
		\noalign{\smallskip}\hline\noalign{\smallskip}
		Time (Second)   & 1.68 & 0.98 &  0.38   \\
		\noalign{\smallskip}\hline
	\end{tabular}
\end{table}

\subsection{Image Processing}\label{sec5a}
This section verifies the proposed SOM-based centerline extraction algorithm and describes the image processing steps.

\begin{figure}
	\centering
	\subfloat[Reference \cite{qi2020adaptive}]{\includegraphics[scale=0.17]{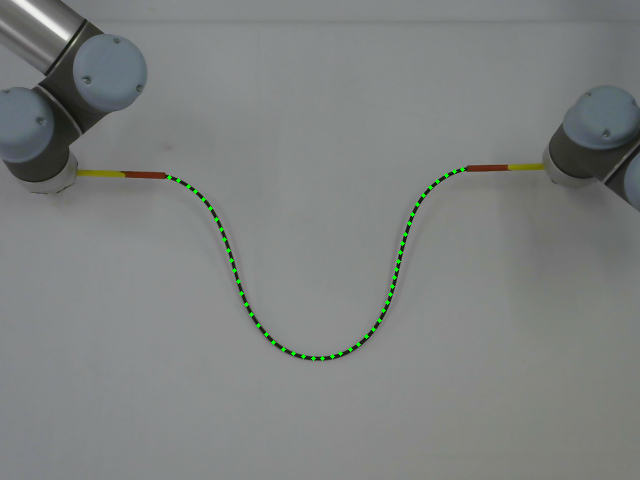}}
	\subfloat[CL \cite{park2000centroid}]{\includegraphics[scale=0.17]{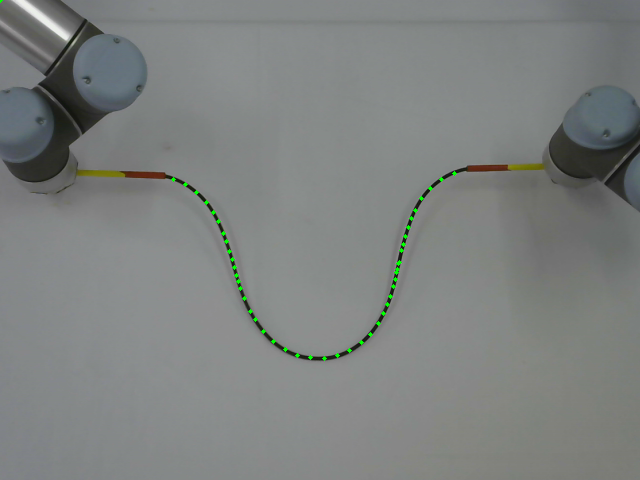}}
	\subfloat[Proposed SOM]{\includegraphics[scale=0.17]{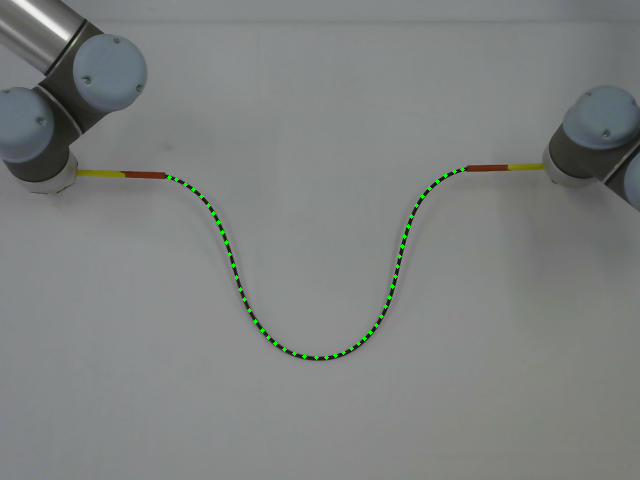}}

	\caption{Comparison of three centerline extraction algorithms, including reference \cite{qi2020adaptive}, CL-based \cite{park2000centroid} and the proposed SOM-based.}
	\label{fig18}
\end{figure}

First, the SOM-based method proposed in this article is compared with two other centerline extraction algorithms.
The first one is based on the \emph{OpenCV/thinning} developed in Reference \cite{qi2020adaptive}, and the second one is based on CL, which is the traditional clustering method \cite{park2000centroid}.
For the fairness of competition, all algorithms need to provide an ordered, fixed-length $N=50$, equidistant centerline.
As the CL-based and SOM-based algorithms only generate an unordered fixed-length centerline, the sorting algorithm \cite{qi2020adaptive} is used to sort unordered centerlines.
For SOM, an open-source toolbox provided by \cite{vettigliminisom} is utilized.
The SOM-based algorithm is the fastest, and it can efficiently perform centerline extraction, as shown in Table \ref{table4}.
One reason is that the provided SOM toolbox is already highly optimized.
Another reason is that the centerline produced by CL-based and SOM-based clustering algorithms has the advantage of fixed-length and equidistant-sampling. 
This method saves more time than \cite{qi2020adaptive}, which sorts all the points and then perform down-sampling.
The proposed SOM-based algorithm and \cite{qi2020adaptive} have similar centerline extraction accuracy, and CL-based has the worst performance, as shown in Fig. \ref{fig18}.
Thus, in terms of accuracy, considering that the processing speed of SOM is the fastest, the SOM-based centerline extraction algorithm is used.

\begin{figure}
	\centering
	\subfloat[ROI selection]{\includegraphics[scale=0.2]{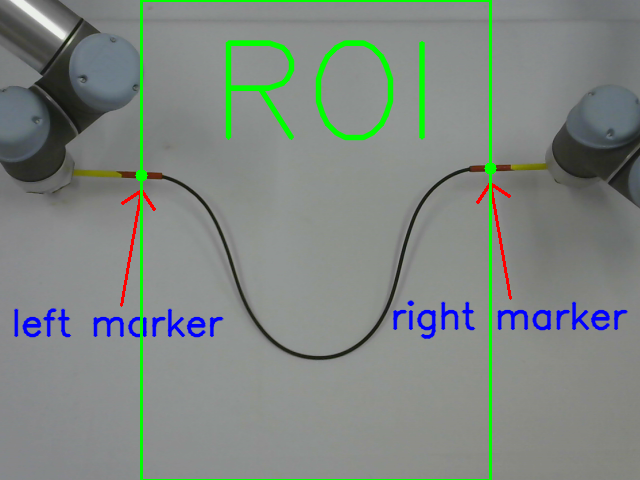}\label{fig18a}}
	\subfloat[Thresholding]{\includegraphics[scale=0.2]{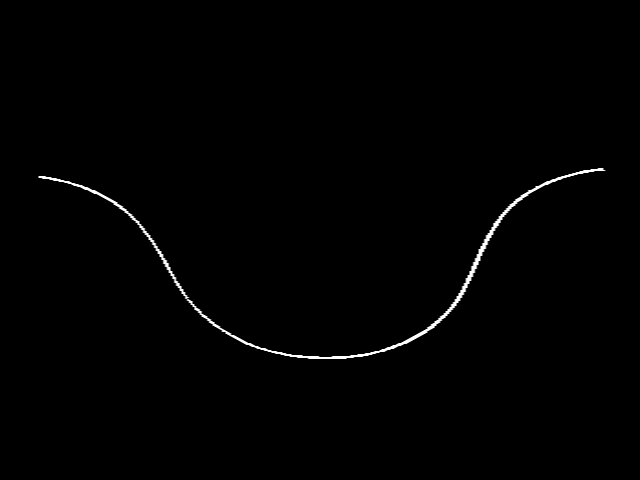}\label{fig18b}}
	
	\subfloat[Centerline extraction (SOM)]{\includegraphics[scale=0.2]{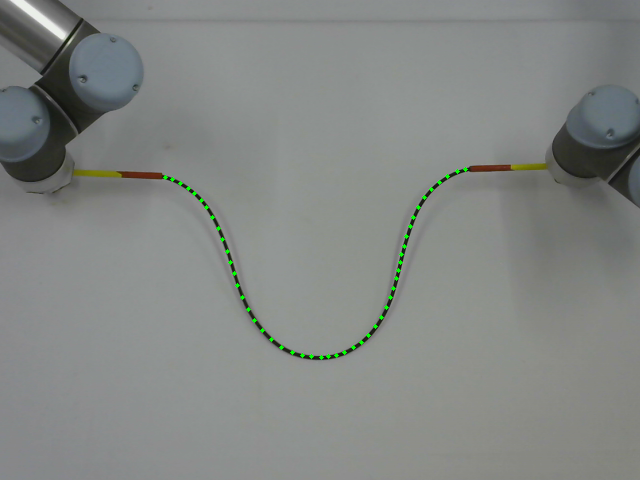}\label{fig18c}}
	\subfloat[Coordinate sorting]{\includegraphics[scale=0.2]{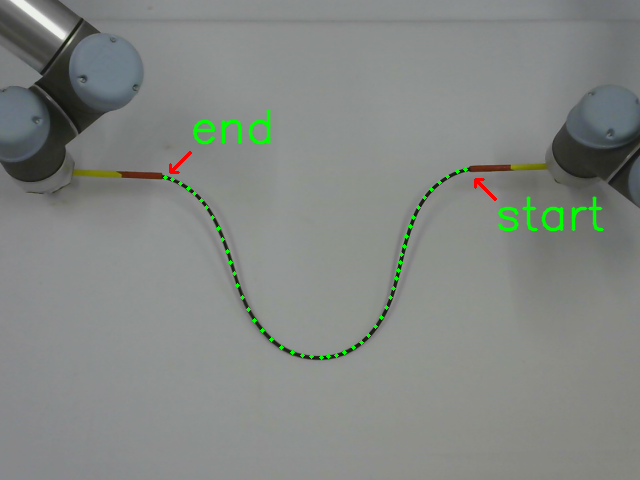}\label{fig18d}}
	
	\caption{Image processing steps.}
	\label{fig19}
\end{figure}

Second, the relevant image processing for centerline extraction is provided. 
The overall process (shown in Fig. \ref{fig19}) is as follows:
\begin{enumerate}
	\item First, segment the red areas nearby the Gripper1 and Gripper2 on the basis of HSV color space and mark them as two green marker points.
	Then, segment the region of the interest (ROI) containing the rod following both green marker points (see Fig. \ref{fig18a}).
	
	\item Next, identify the rod in ROI, remove the noise, and obtain a binary image of the rod using OpenCV morphological opening algorithm (see Fig. \ref{fig18b}).
	
	\item Subsequently, use the proposed SOM-based algorithm to get an unordered centerline with $N = 50$ (see Fig. \ref{fig18c}).
	
	\item Finally, apply the sorting algorithm \cite{qi2020adaptive} to get an ordered centerline (see Fig. \ref{fig18d}). 
	The starting point is the closest point to the right marker point on the centerline.
\end{enumerate}

\begin{figure}
	\centering
	\includegraphics[scale=0.35]{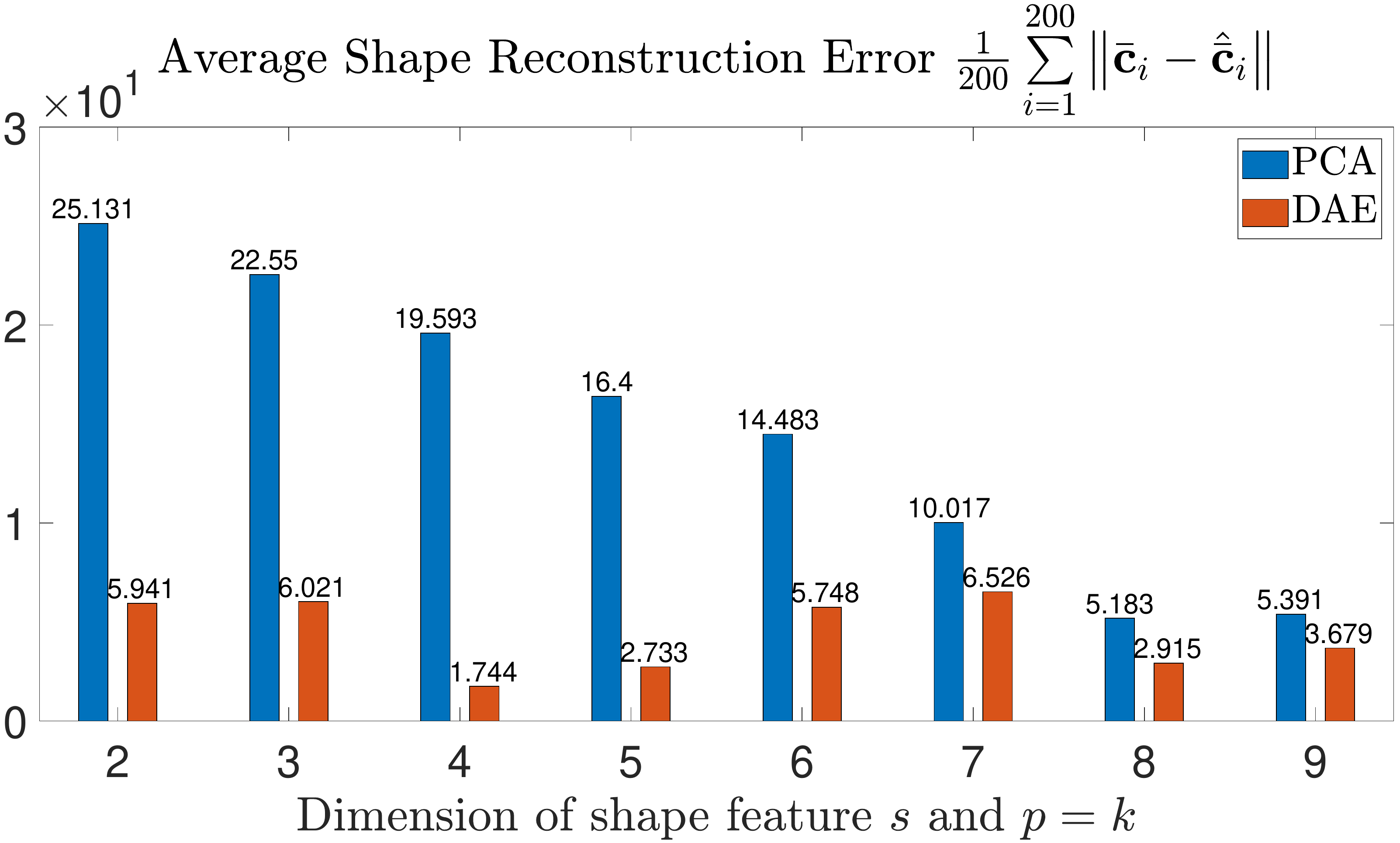}
	\caption{Average shape reconstruction error comparison between DAE and PCA among 200 shape sets in the experiment.}
	\label{fig7}
\end{figure}

\subsection{Feature Extraction Comparison}
Similar to Section \ref{sec4a}, 40,000 samples are generated in the same way.
The structure of DAE is similar with Section \ref{sec4a}, as shown in Fig. \ref{fig30}.
Batch-Normalization-1D (BN) is added after each layer.
From the comparison results shown in Fig. \ref{fig7}, the reconstruction accuracy of DAE is still better than PCA.
For DAE, the accuracy of ${p}=4$ is the best, and $p=5$ is the second best.
This finding is consistent with the simulation results.
The results prove the effectiveness of the proposed DAE in the shape feature extraction.
Thus DAE with $p=4$ is used in the following sections.

\begin{figure}
	\centering
	\subfloat[]{\includegraphics[scale=0.2]{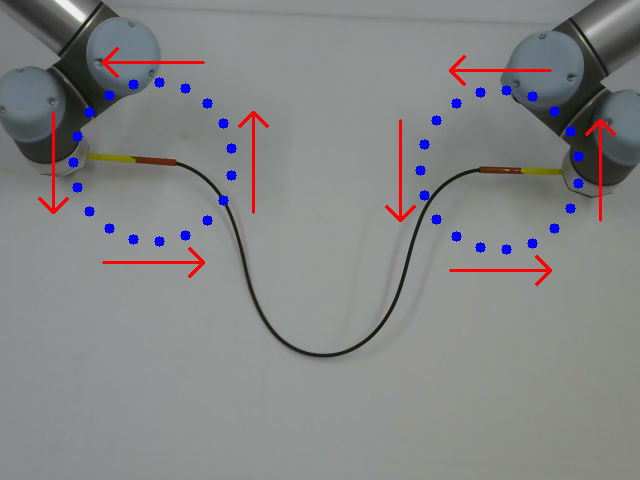}\label{fig23a}}
	\subfloat[]{\includegraphics[scale=0.2]{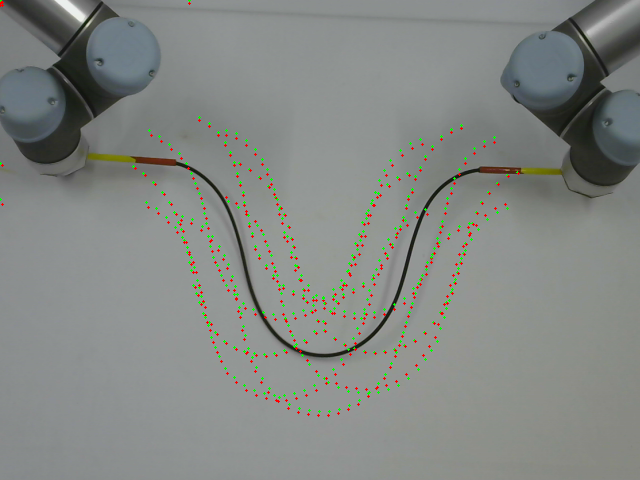}\label{fig12b}}
	
	\caption{Deformation Jacobian matrix $\hat{\mathbf{J}}_k$ validation framework. 
	(a) Motion trajectory of robot's end-effector.
	(b) Comparison between the visually measured cable profile (green dot line) and its reconstruction shape obtained by DAE (red dot line) with $p=4$.}
	\label{fig25}
\end{figure}

\begin{figure}
	\centering
	\includegraphics[scale=0.3]{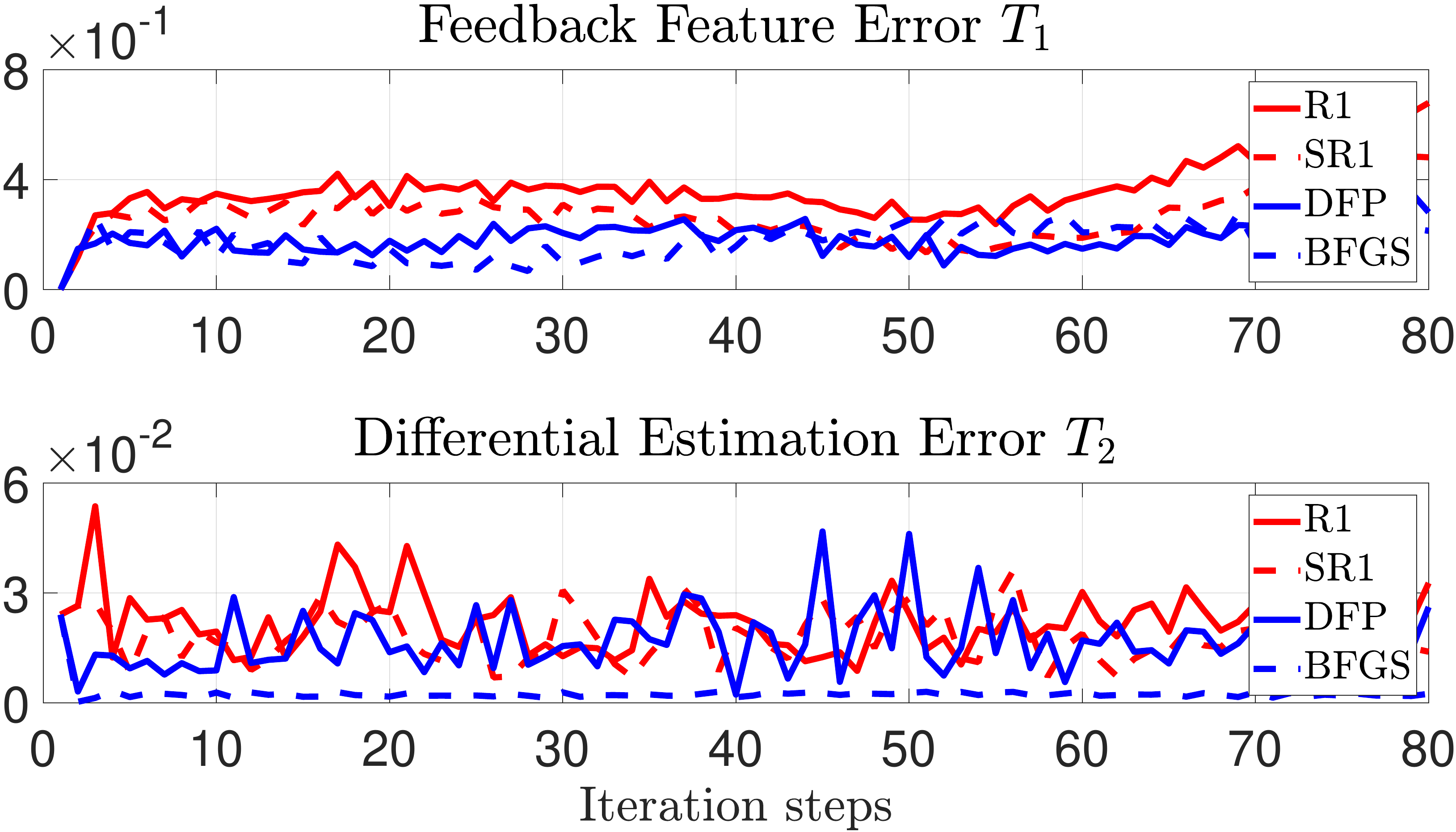}
	\caption{Profiles of the criteria $T_1$ and $T_2$ that are computed along the circular trajectory.}
	\label{fig11}
\end{figure}

\subsection{Validation of the Jacobian Estimation}
Similar to Section \ref{sec4b}, two UR5 are commanded to move along a fixed circular trajectory, as depicted in Fig. \ref{fig23a}.
The shape reconstruction performance of DAE is accurate in the experiment, as shown in Fig. \ref{fig12b}.
Thus, whether for simulation or experiment, DAE can be applied in the shape representation.
The comparison results depicted in Fig. \ref{fig11} show that BFGS has a minimal deformation Jacobian approximation error, which validates its effectiveness and adaptability in different regions.

\subsection{Manipulation of Elastic Rods}
Similar to Section \ref{sec4c}, a desired shape $\bar{\mathbf{c}}^*$ should be given in advance.
The following steps are conducted to obtain the feasible target shape.
\begin{itemize}
    \item First, the robot is moved to a position while avoiding the singular shapes, e.g., straight.
    
    \item Second, the current shape of the elastic rod is recorded by the camera and denoted by $\bar{\mathbf{c}}^*$ as the target shape.
    
    \item Third, the target shape $\bar{\mathbf{c}}^*$ is fed into the trained DAE to get target shape feature denoted by $\mathbf{s}^*$.
    
    \item Fourth, the robot moves back to the initial position and starts the deformation with the given $\mathbf{s}^*$.
\end{itemize}

Considering safety, the saturation of $\Delta \mathbf{r}$ is set to, $\left| {\Delta {r_{i1}}} \right| \le 0.01m/s$, $\left| {\Delta {r_{i2}}} \right| \le 0.01m/s$ and $\left| {\Delta {r_{i3}}} \right| \le 0.1rad/s, i=1,2,3$, respectively.
Four experiments with different initial and target shapes are conducted to verify the proposed algorithm's effectiveness.

\begin{figure}
	\centering
	\subfloat[Experiment1-R1]{\includegraphics[scale=0.134]{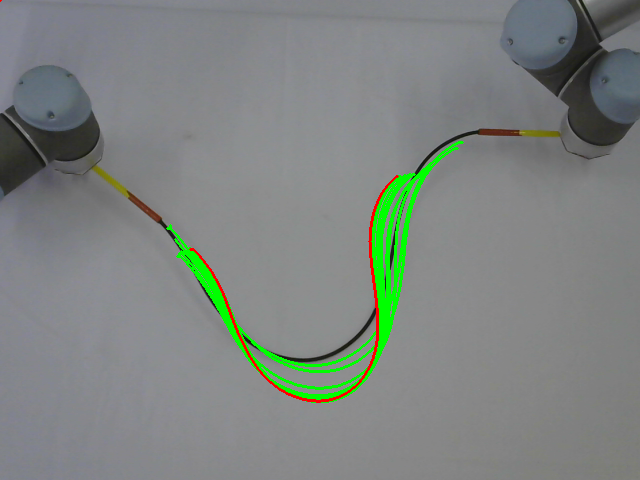}}
	\subfloat[Experiment2-R1]{\includegraphics[scale=0.134]{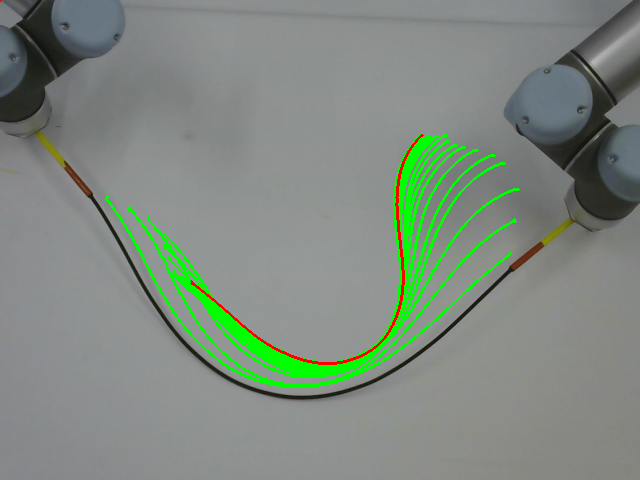}}
	\subfloat[Experiment3-R1]{\includegraphics[scale=0.134]{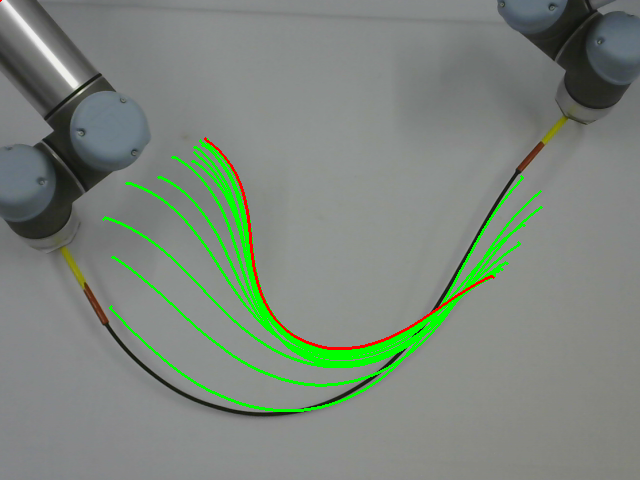}}
	\subfloat[Experiment4-R1]{\includegraphics[scale=0.134]{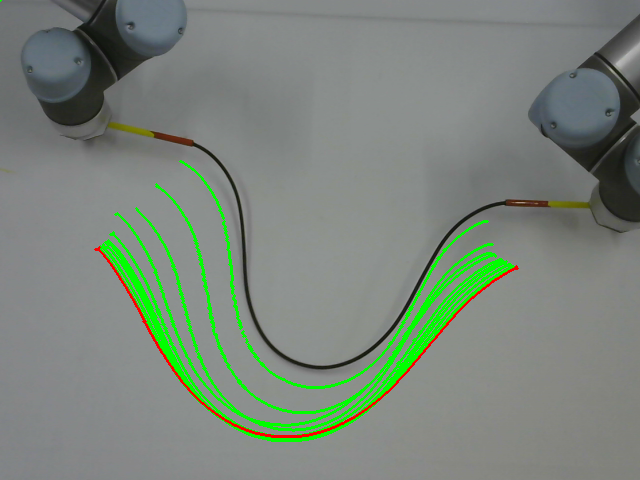}}

	\subfloat[Experiment1-SR1]{\includegraphics[scale=0.134]{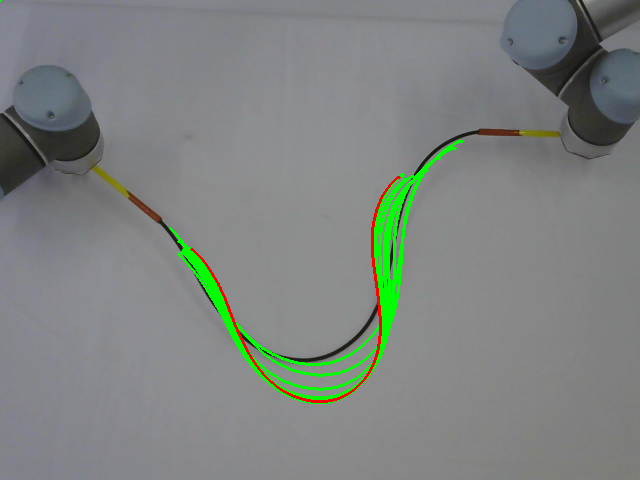}}
	\subfloat[Experiment2-SR1]{\includegraphics[scale=0.134]{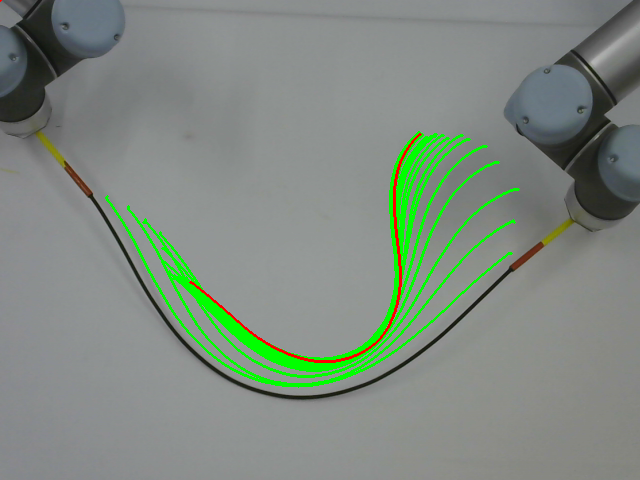}}
	\subfloat[Experiment3-SR1]{\includegraphics[scale=0.134]{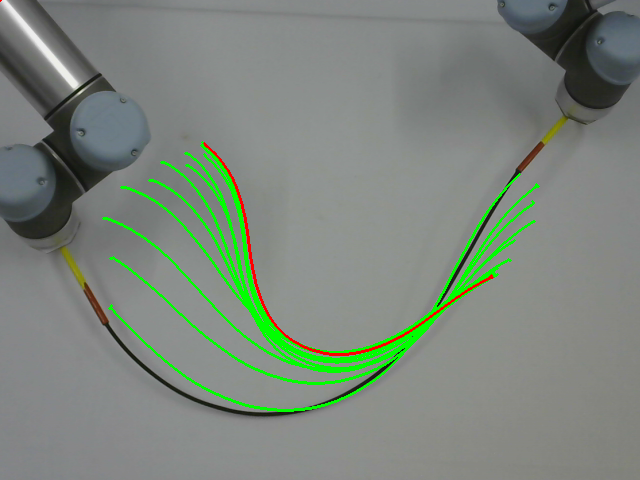}}
	\subfloat[Experiment4-SR1]{\includegraphics[scale=0.134]{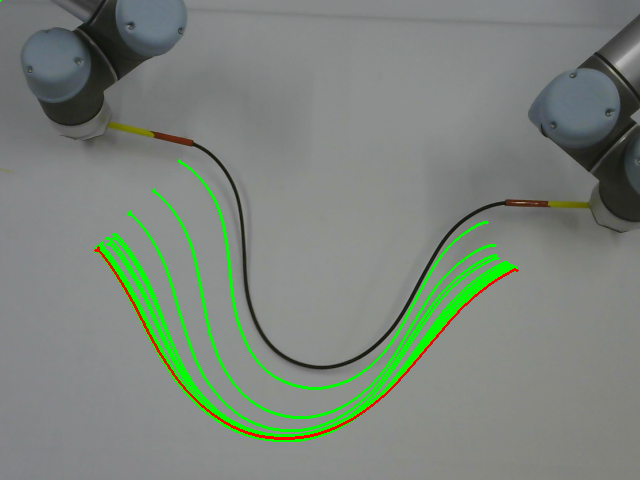}}

	\subfloat[Experiment1-DFP]{\includegraphics[scale=0.134]{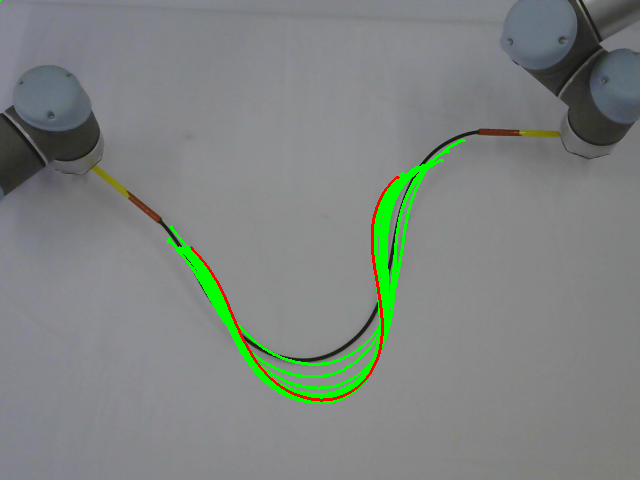}}
	\subfloat[Experiment2-DFP]{\includegraphics[scale=0.134]{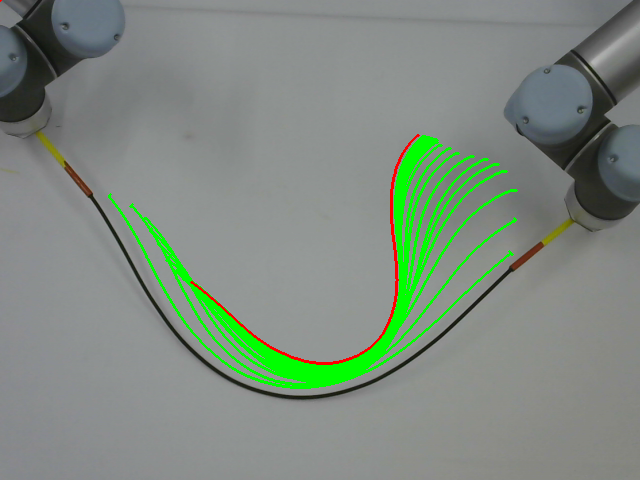}}
	\subfloat[Experiment3-DFP]{\includegraphics[scale=0.134]{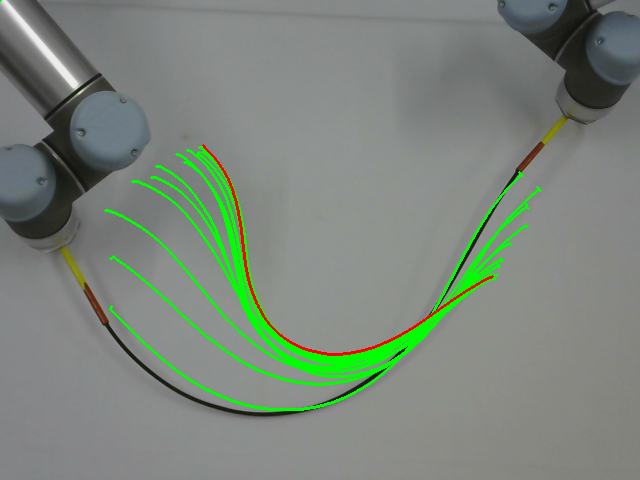}}
	\subfloat[Experiment4-DFP]{\includegraphics[scale=0.134]{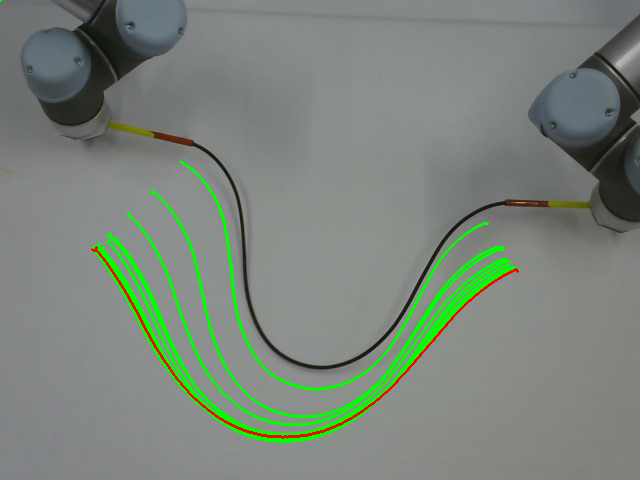}}

	\subfloat[Experiment1-BFGS]{\includegraphics[scale=0.134]{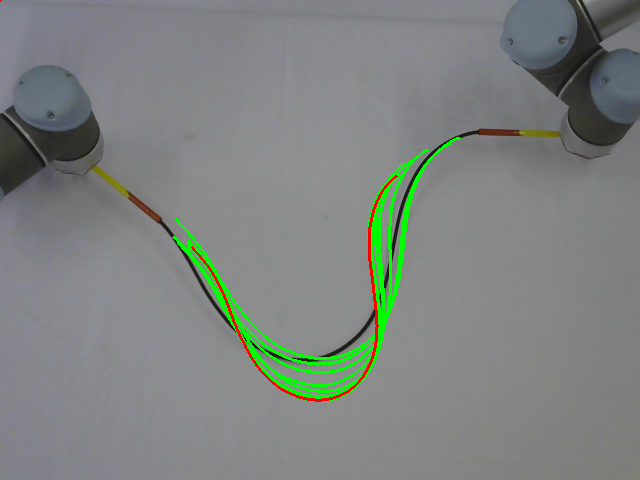}}
	\subfloat[Experiment2-BFGS]{\includegraphics[scale=0.134]{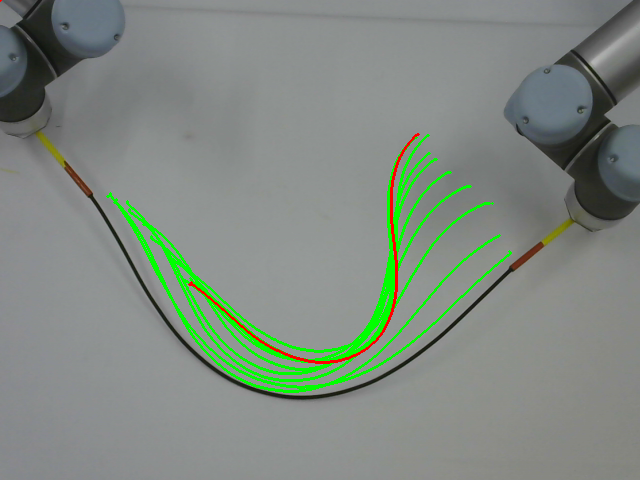}}
	\subfloat[Experiment3-BFGS]{\includegraphics[scale=0.134]{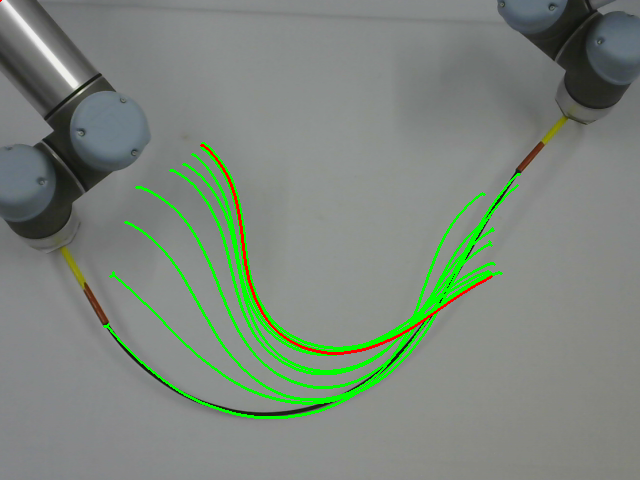}}
	\subfloat[Experiment4-BFGS]{\includegraphics[scale=0.134]{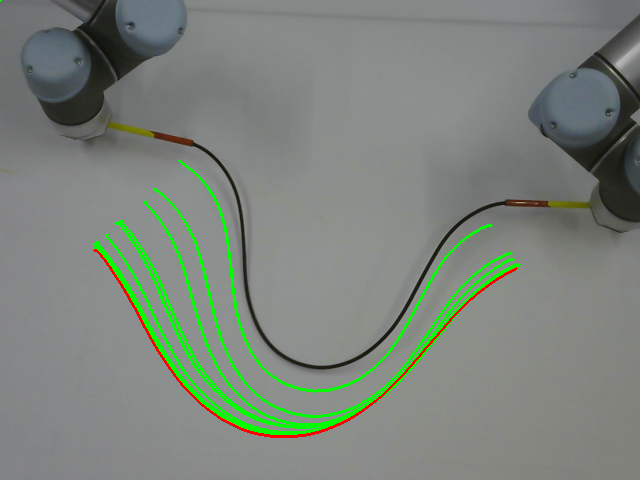}}

	\caption{Initial (black solid line), transition (green solid line) and target (red solid line) configurations in the four shape deformation experiments which have a variety of different initial and target shape with dual-UR5 robot among R1, SR1, DFP and BFGS.}
	\label{fig8}
\end{figure}

\begin{figure}
	\centering
	\subfloat[Experiment1 result]{\includegraphics[scale=0.18]{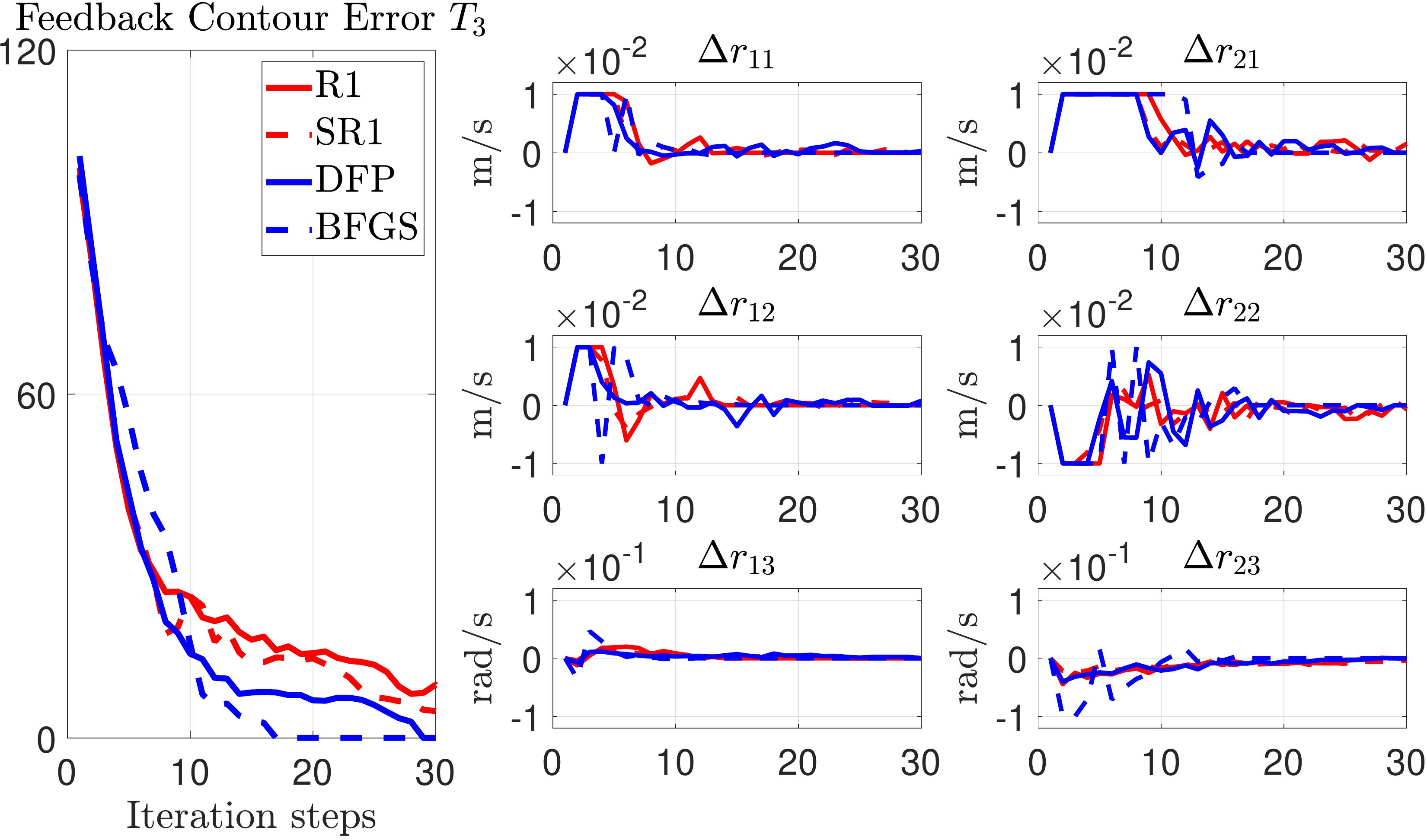}}
	\subfloat[Experiment2 result]{\includegraphics[scale=0.18]{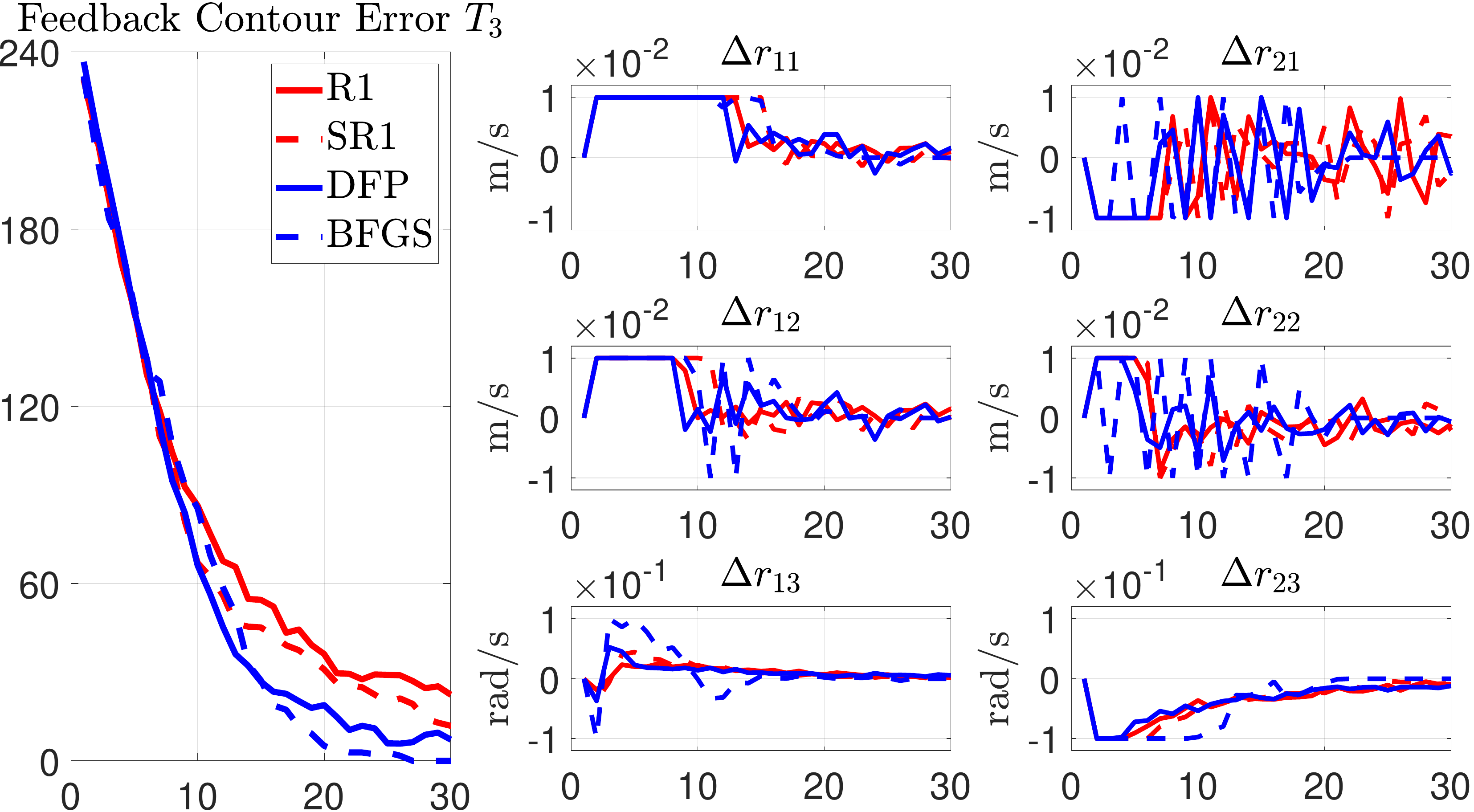}}
	
	\subfloat[Experiment3 result]{\includegraphics[scale=0.18]{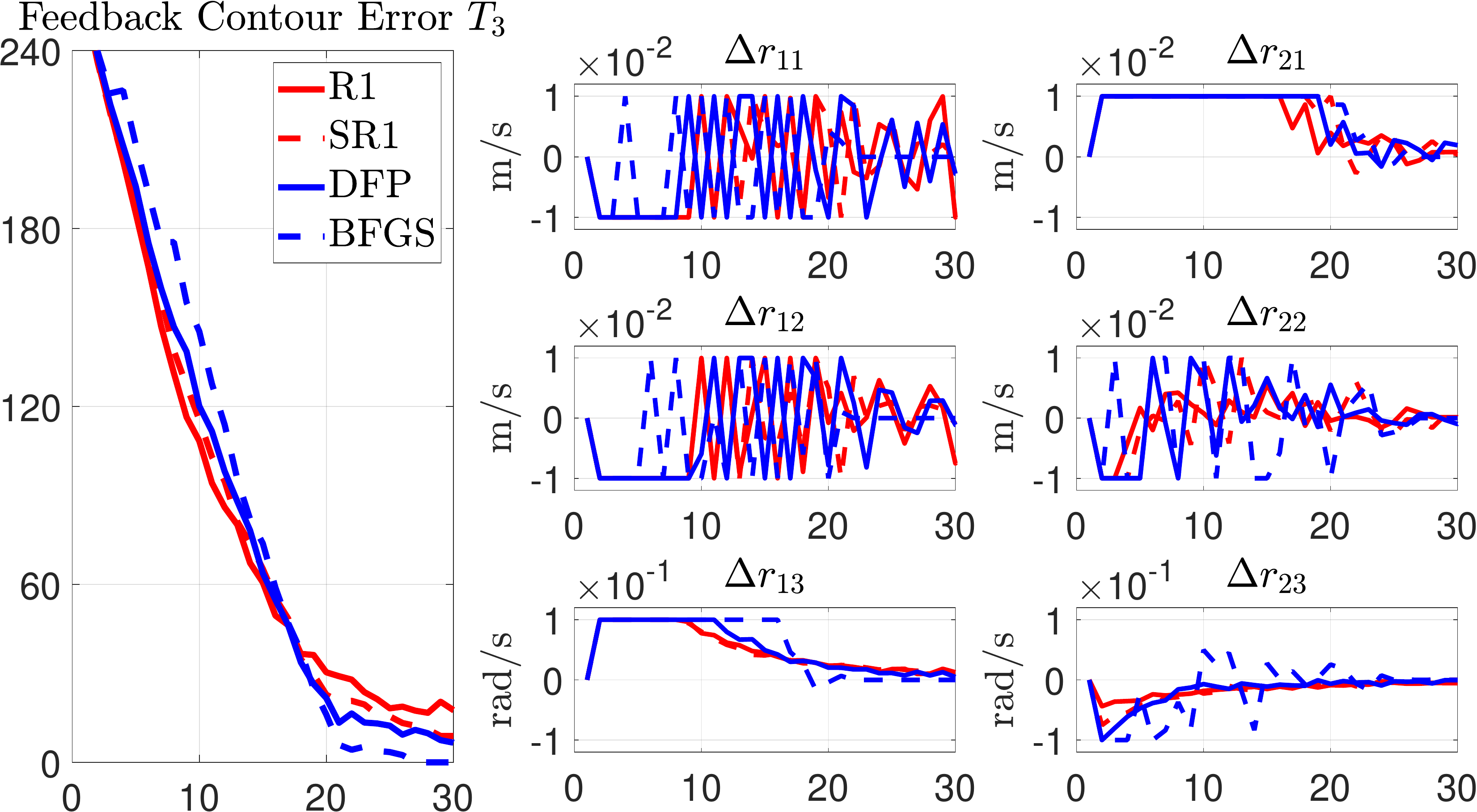}}
	\subfloat[Experiment4 result]{\includegraphics[scale=0.18]{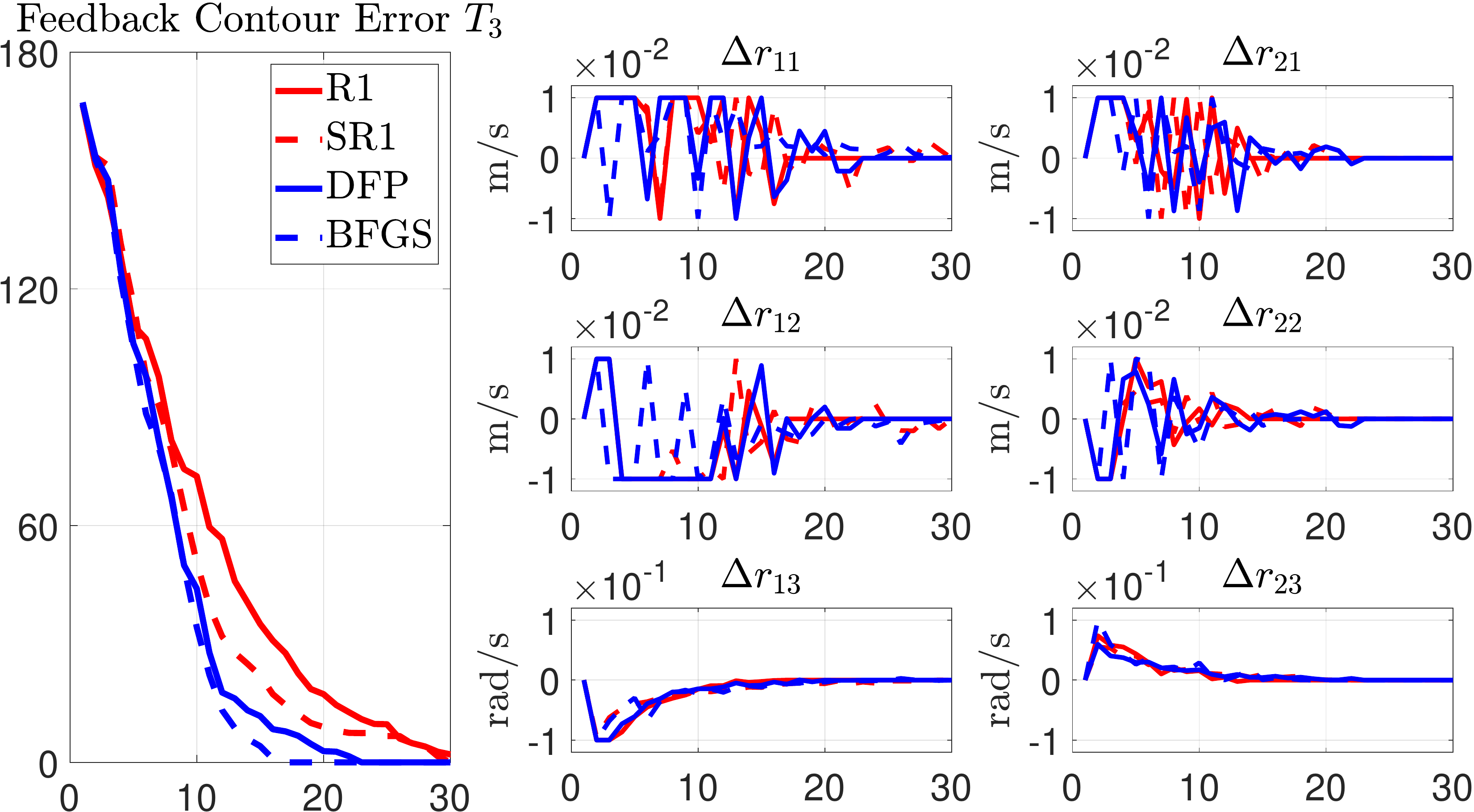}}
	\caption{Profiles of the criterion $T_3$ and velocity command $\Delta \mathbf{r}_k$ among R1, SR1, DFP and BFGS within four shape deformation experiments.}
	\label{fig9}
\end{figure}

In the proposed algorithm, the elastic rod can be deformed using two UR5 to the desired shape accurately without damaging the object during the deformation process, as depicted in Fig \ref{fig8}.
The profiles of $T_3$ and the velocity command $\Delta \mathbf{r}_k$ are shown in Fig. \ref{fig9}.
Corresponding to the simulation results, BFGS still has the most significant control effect, the fastest convergence speed, without apparent fluctuations (large instantaneous deformation).
Thus, BFGS has excellent adaptability and robustness to various conditions in the shape deformation issue.
In this study, the experiment uses two UR5.
Overall, the results prove the effectiveness of the proposed algorithm for the multi-manipulator shape deformation issue.

\section{Conclusions}\label{sec6}
A framework for the deformation control of elastic rods is proposed without any prior physical knowledge.
It includes shape feature extraction, deformation Jacobian matrix estimation, and a robust SOM-based centerline extraction algorithm.
First, new shape features based on DAE are utilized to represent the elastic rod's centerline in the low-dimensional latent space.
Second, the performance of four deformation Jacobian estimators (R1, SR1, DFP, and BFGS) is evaluated.
Third, the velocity controller is derived and the system stability is proven.
Finally, the effectiveness and feasibility of the proposed algorithm are validated by the numerical and experimental results.

DAE is used in this study to map the high-dimensional geometric information of elastic rods flexibly into a low-dimension latent space.
The proposed feature extraction algorithm has better shape representation capabilities than the traditional PCA.
It also does not require any artificial markers, making it widely applicable to practical situations.
Broyden algorithms are used to approximate the deformation Jacobian matrix in real-time.
In this way, the physical parameters and camera models are not identified.
From the results, BFGS has the advantages of simple structure, fast calculation speed, and accurate approximation performance.
Simultaneously, a robust SOM-based centerline extraction algorithm with a fast calculation speed and high extraction accuracy is designed.
The overall system is completely calculated from the visual feedback data, without any prior physical characteristics of the elastic rod and the requirement to calibrate the camera.

The proposed method also has some limitations.
First, the manipulated object is only soft elastic objects, e.g., carbon fiber rod. 
Thus the proposed algorithm is not suitable for inelastic items, e.g., plasticine and rope.
Second, although DAE has a good shape representation ability, it needs an extensive and rich-enough dataset to train itself, which has particular difficulties in practical applications.
Third, the approximation of deformation Jacobian matrix based on Broyden algorithms is easy to fall into the local optimum, which may generate the destructive operation, such as over-tension and over-compression in the manipulation process. 

In the future, 3D deformation tasks will be included to manipulate more complex shapes, e.g., M-shaped and spiral.
Moreover, the existing DAE needs to be improved to be suitable for different scenarios and materials.
Path planning should be considered to avoid possible destructive operations during the manipulation process.

\begin{acknowledgements}
This work was supported in part by the Germany/Hong Kong Joint Research
Scheme sponsored by the Research Grants Council of Hong Kong and the German Academic Exchange Service under grant G-PolyU507/18, in part by the Research Grants Council of Hong Kong under grant number 14203917, in part by the Key-Area Research and Development Program of Guangdong Province 2020 under project 76.
\end{acknowledgements}

%
%

\bibliographystyle{spmpsci}      
\bibliography{david_biblio.bib}


\end{document}